\documentclass[runningheads]{llncs}

\usepackage{eccv}

\usepackage{eccvabbrv}

\usepackage{graphicx}
\usepackage{booktabs}
\usepackage{makecell}
\usepackage{wasysym}
\usepackage{pifont}
\usepackage{multirow}
\usepackage{marvosym}

\usepackage[accsupp]{axessibility}  

\usepackage{hyperref}

\usepackage{orcidlink}
\usepackage{algorithm}
\usepackage{algorithmic}
\usepackage{graphicx}
\usepackage{amsmath}
\usepackage[table]{xcolor}
\usepackage{appendix}

\usepackage{hyperref} 
\usepackage{fontawesome} 

\begin{document}

\title{Illuminating Unified Multimodal Model for Free-form Interleaved Text-Image Generation}

\titlerunning{ILLUME-X}

\author{
Chonghuinan Wang\inst{1,*} 
\and
Zhikai Chen\inst{2,*}
\and
Chunwei Wang\inst{2,*}
\and
Yecong Wan\inst{1,3}
\and
Junwei Yang\inst{2}
\and
Zhixin Wang\inst{2}
\and
Wei Zhang\inst{2}
\and
Jiaqi Xu\inst{2}
\and
Renjing Pei\inst{2}
\and \\
Xiaohe Wu\inst{1,}\textsuperscript{\Letter}
\and
Fan Li\inst{2,4,\dag,}\textsuperscript{\Letter}
\and
Wangmeng Zuo\inst{1}
}

\authorrunning{C.~Wang et al.}

\institute{Harbin Institute of Technology, Harbin, China 
\and
Huawei Noah’s Ark Lab, Shenzhen, China 
\and
Zhengzhou Advanced Research Institute of Harbin Institute of Technology, Zhengzhou, China
\and
Nankai University, Tianjin, China
\\
\email{25b903050@stu.hit.edu.cn, csxhwu@gmail.com, lifan61@huawei.com}\\
\faGithub \space \url{https://github.com/ChonghuinanWang/ILLUME-X} 
}
\maketitle

\begin{figure}
    \centering
    \includegraphics[width=\linewidth]{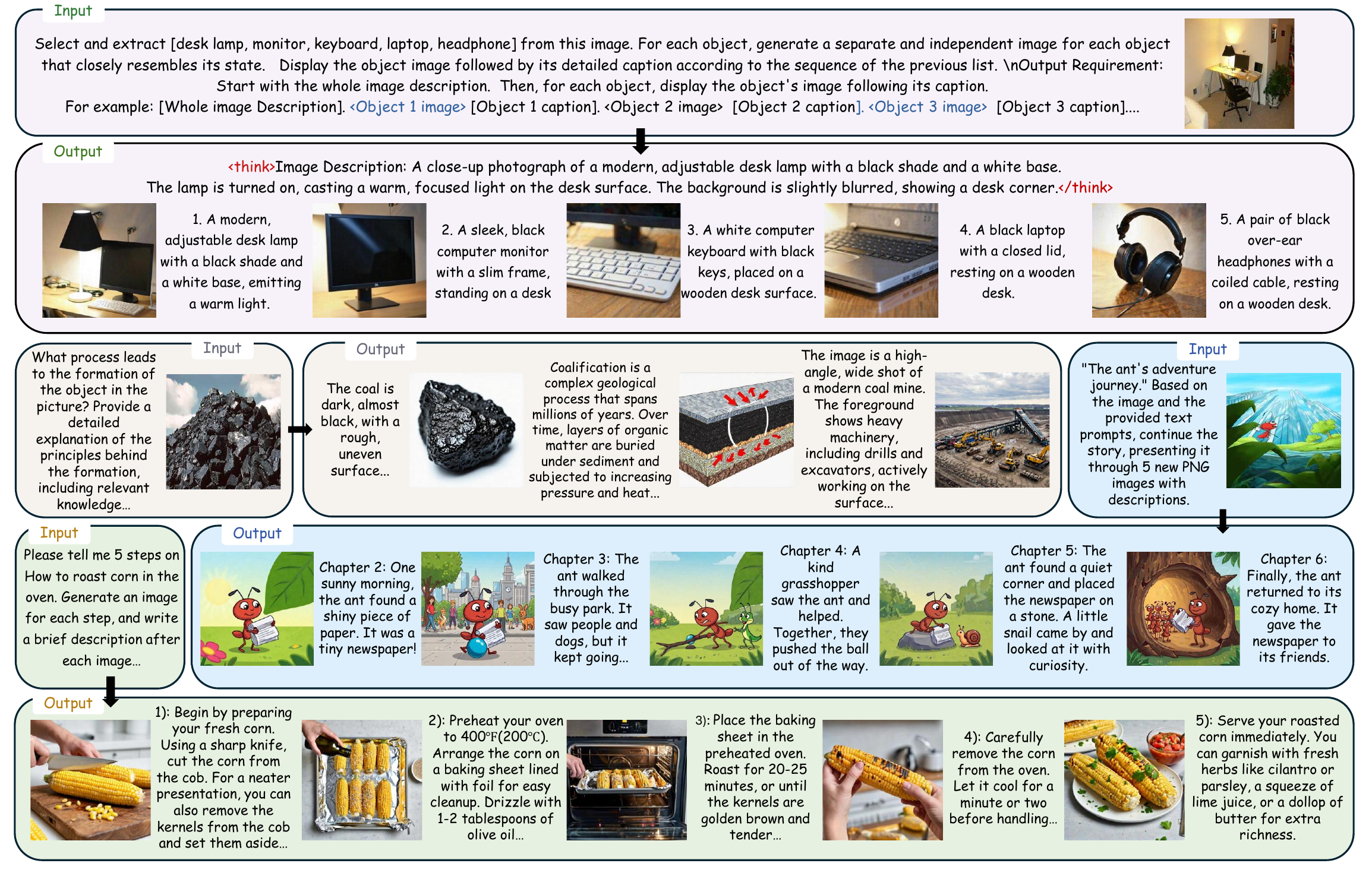}
    \caption{
    \textbf{Illustrative examples of ILLUME-X.} The model handles interleaved text-image inputs and outputs, enabling cohesive multimodal understanding and generation.
    }
    \label{fig:placeholder}
\end{figure}

\renewcommand{\thefootnote}{}
\footnotetext{* Equal Contribution, \space \Letter \space Corresponding Author, \space \dag \space Project Leader}

\begin{abstract}
The advancement of generative AI models capable of producing text and image marks a critical step forward in the realm of multimodal intelligence, particularly for tasks involving the interleaving of both modalities. To advance this intelligence to the next stage, it is crucial for models to autonomously generate free-form interleaved text-image sequences.
In this paper, we introduce \textbf{ILLUME-X}, an advanced unified multimodal paradigm that enables high-quality, free-form interleaved text-image generation by improving multimodal data efficiency and stabilizing the multimodal training process. 
ILLUME-X comprises three key components: (i) an expanded training data pipeline optimized for interleaved text-image generation, (ii) a progressive training strategy with self-adaptive objectives for free-length multimodal token sequences, and (iii) an objective and comprehensive evaluation method \textbf{ILScore} for interleaved text-image sequences.
Notably, our ILLUME-X outperforms previous unified models across multiple interleaved text-image generation tasks like style transfer, image decomposition and storytelling.
\end{abstract}

\section{Introduction}
\label{sec:intro}

Multimodal Large Language Models (MLLMs) and diffusion models \cite{liu2024smartcontrol, chameleon, transfusion, janus, januspro, janusflow, bagel, emu35, illumeplus, emma, Wang_2026_CVPR_CREval} have recently achieved significant advances in visual understanding and generation. By merging understanding and generation capabilities within a single framework, unified multimodal models can genuinely grasp the relationships between visual and textual information to enable more intelligent interactions and task execution. 
Naturally, researchers demonstrate considerable interest in their abilities for interleaved text-image generation, which requires the models to generate both images and text pieces in an arbitrary sequence for diverse real-world scenarios. 

To enable multimodal models to perform interleaved generation, existing studies have explored several complementary paradigms. One line of work~\cite{dreamllm, emu35, nextflow} directly applies a pure auto-regression model to generate texts and images. These models utilize a unified auto-regressive Transformer to learn the joint probability distribution of text and visual tokens, and elicit emergent cross-modal generative capabilities by scaling up dataset size and model capacity.
Another direction \cite{mminterleaved, orthus} attempts to combine the diffusion model into an auto-regression model. These approaches incorporate modality-aware architectural designs within the transformer to support joint prediction of text and image tokens, thereby enabling cross-modal generation in a single framework.
However, despite these advancements, existing methods have not fully evaluated the performance of their models or sufficiently validated their capabilities on interleaved text-image generation benchmarks. This gap in comprehensive evaluation and benchmark performance is a key motivation for this work.

In this work, we propose ILLUME-X, a unified multimodal paradigm that enables high-quality, free-form interleaved text-image generation. 
To empower the model with high-quality interleaved capabilities, we first establish a systematic data pipeline that curates 100K high-fidelity training samples. This pipeline integrates authentic interleaved sequences derived from video data with sophisticated synthetic data generated by multimodal LLMs, capturing real-world temporal dynamics and diverse generative scenarios.
Building upon this data foundation, ILLUME-X employs an integrated architecture that aligns cross-modal representations using shared attention mechanisms. Additionally, an interleaved training paradigm is introduced, featuring a specialized attention mask to jointly optimize text generation, image synthesis, and modality transitions. This enables flexible multi-input-to-multi-output (N-to-M) reasoning.
Finally, we propose a rigorous evaluation protocol ILScore to address the lack of task-specific metrics for interleaved generation. This framework evaluates the generated sequences from two critical perspectives: cross-modal continuity (ensuring seamless integration) and unimodal generative quality (ensuring both textual coherence and visual fidelity).

The main contributions of ILLUME-X are as follows:
\begin{itemize}
\item It introduces a pioneering unified multimodal paradigm that achieves free-form N-to-M interleaved generation, powered by a progressive training strategy with self-adaptive objectives for free-length multimodal token sequences.
\item It presents an efficient and scalable data pipeline that curates 100K high-quality interleaved samples, enhancing model training.
\item It establishes a multi-dimensional evaluation protocol ILScore specifically designed for complex text-image sequences, ensuring thorough performance assessment across modalities.
\item Extensive experiments demonstrate that ILLUME-X outperforms previous methods, enabling unified MLLMs to generate free-form interleaved content.
\end{itemize}

\section{Related Work}
\subsection{Unified Multimodal Models}
Unified multimodal models have emerged as a key direction for bridging the gap between multimodal understanding and generation, aiming to consolidate diverse tasks (\eg vision-language reasoning, text-to-image synthesis, and image editing) within a single architectural framework. Existing unified models can be broadly grouped into three main categories: diffusion models, autoregressive (AR) models, and hybrid combinations of the two paradigms. 
Diffusion-based unified models (\eg Dual Diffusion~\cite{li2025dual}, MMaDA~\cite{yang2025mmada}, LaViDa-O~\cite{li2025lavida}, UniModel~\cite{zhang2025unimodel}, Muddit~\cite{shi2025muddit}) usually adapt text-to-image diffusion pipelines to multimodal scenarios by conditioning the denoising process on cross-modal contextual signals. They offer high visual fidelity but are limited by slow inference and weaker sequential reasoning.
AR-based unified models (\eg Chameleon~\cite{chameleon}, Janus-Pro~\cite{januspro}, UniToken~\cite{jiao2025unitoken}, Emu 3~\cite{wang2024emu3}) typically serialize vision and language tokens into a single sequence and model them in an ordered, token-by-token manner.
AR + diffusion hybrid models (\eg Transfusion~\cite{transfusion}, Show-o~\cite{xie2024show-o},  BLIP3-o~\cite{chen2025blip3},  OmniGen2~\cite{omnigen2}, Mogao~\cite{mogao}, EMMA~\cite{emma}, BAGEL~\cite{bagel}, HBridge~\cite{wang2025hbridge}, HunyuanImage~\cite{cao2025hunyuanimage}) offer a compelling compromise, combining the compositional control of autoregressive decoding with the high-fidelity synthesis of diffusion.

\subsection{Interleaved Generation Models and Datasets}
Although existing unified models can produce both images and text, they are typically constrained to a single target modality specified in advance. While some studies (\eg~\cite{bagel, xie2025show-o2, mogao, emu35, cao2025hunyuanimage, Li_2026_CVPR, Wu_2026_CVPR_YOSE}) appear to demonstrate the feasibility of interleaved image–text generation, dedicated investigation and validation for N-to-M interleaved generation scenarios remain limited. To this end, CoMM~\cite{chen2025comm} aggregates multi-source data with an emphasis on instructional content and visual storytelling to strengthen MLLMs’ in-context learning and interleaved image–text generation. In parallel, Zebra-CoT~\cite{li2025zebra} provides large-scale interleaved multimodal reasoning data, and fine-tuning BAGEL~\cite{bagel} on it equips the model with interleaved multimodal reasoning capabilities. However, despite their utility, both datasets remain limited in overall quality and consistency, which may be hinder stable training for N-to-M interleaved generation. Furthermore, frameworks such as WEAVE~\cite{chow2025weave} and ISG-Bench~\cite{chen2024interleaved-isg} provide a critical foundation for evaluating interleaved generation by establishing standardized task settings and unified evaluation protocols.

\section{Approach}

\begin{figure}[t]
    \centering
    \includegraphics[width=\textwidth]{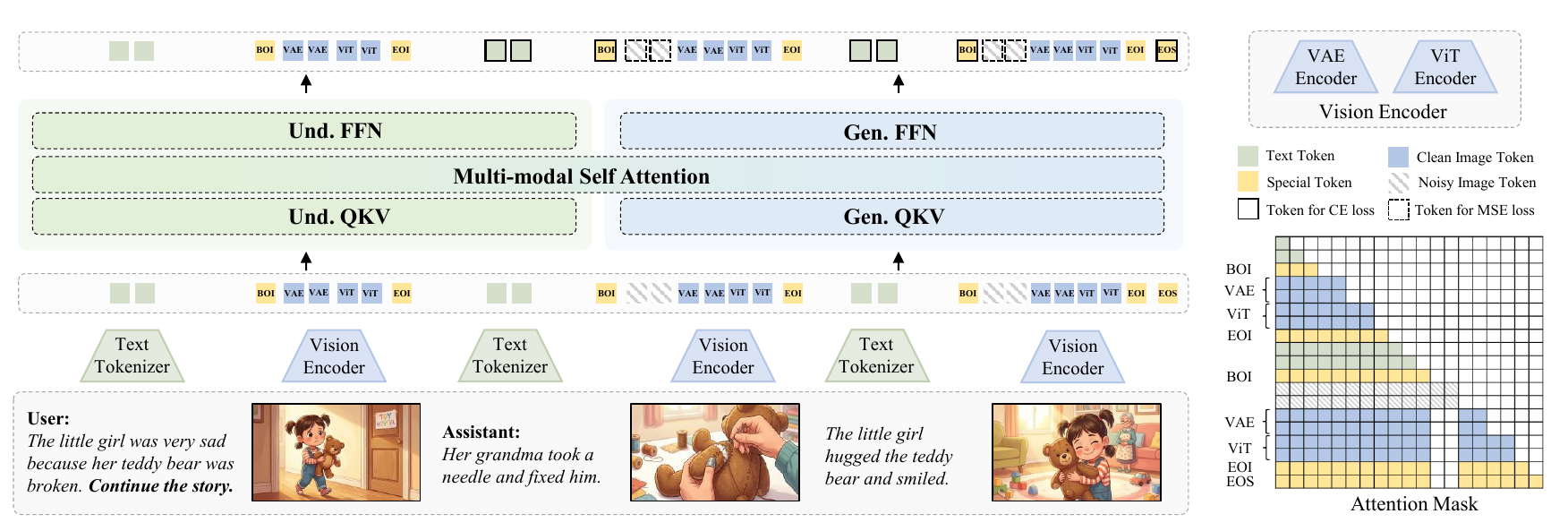}
    \caption{\textbf{The overall architecture of our ILLUME-X.} ``QKV'' and ``FFN'' represent query, key, value vectors for attention computation and feedforward networks, respectively.} 
    \label{fig:model}
\end{figure}

\subsection{Model Architecture}
\qquad\textbf{Unified Transformer.}
The architecture of ILLUME-X is illustrated in Figure~\ref{fig:model}. Following established practices~\cite{janus,januspro}, we employ distinct continuous visual encoders to support both visual understanding and generation tasks. Specifically, a ViT encoder is used to extract visual semantic features, while a VAE encoder captures low-level visual details. 
The backbone model adopts a decoder-only transformer architecture, consistent with modern large language models~\cite{qwen25vl,qwen3vl}. Following the MoT~\cite {bagel} paradigm, our transformer processes token sequences via shared self-attention across all layers, and employs modality-specific QKV projections and FFNs to support both image understanding and generation.
To ensure stable and efficient training, the model integrates several advanced components, including RMSNorm~\cite{rmsnorm} for normalization, QK-Norm~\cite{scalevit,scaleflowmatching} to stabilize attention gradients, SwiGLU~\cite{swiglu} as the activation function, and RoPE~\cite{rope} for token positional encoding. Additionally, we utilize Grouped-Query Attention~\cite{gqa} to reduce KV cache overhead and improve inference efficiency. 

\textbf{Interleaved Attention.} To handle free-form interleaved multimodal contexts, our framework concatenates text tokens, vision tokens and special tokens into a unified sequence. 
Figure~\ref{fig:model} illustrates the token organization mechanism and their attention mask in our model. Besides the tokens encoded from text and image, we introduce special tokens, including BOI (begin-of-image), EOI (end-of-image), and EOS (end-of-sequence) tokens, to explicitly demarcate modality transitions. 
Consequently, text and vision tokens from both understanding and generation tasks are interleaved within the model according to the structural layout of the input. For tokens within the same sample, we employ a generalized causal attention mechanism. Specifically, tokens are partitioned into consecutive segments based on their modality. While tokens in any given segment can attend to all tokens in preceding segments, the attention within each segment is task-specific: we apply causal attention to text and special tokens to maintain the auto-regressive property, while utilizing bidirectional attention for vision tokens to capture holistic spatial information. 
We explicitly arrange the noisy latent tokens and clean reconstruction tokens within the VAE latent sequence. To alleviate the error accumulation caused by noise interference, we apply a masking operation to the noisy image tokens during the generation of the next set of samples.

\textbf{Training Objective.} 
During training, we employ a multi-objective optimization strategy to jointly supervise the interleaved sequences. Specifically, a Cross-Entropy (CE) loss is applied to text and special control tokens (BOI/EOS) to maintain language modeling proficiency and ensure seamless modality transitions. Concurrently, vision tokens are optimized via a Rectified Flow-based mean squared error (MSE) loss~\cite{flowmatching,scaleflowmatching}, enabling high-fidelity image synthesis within the unified multimodal stream.

\subsection{Interleaved Classifier-free Guidance}
\label{section32}
For semantic alignment control in image generation, we use Classifier-Free Guidance (CFG) to enhance generation quality, which combines conditional predictions with unconditional predictions to yield results that more closely align with the specified conditions, formulated as:
\begin{equation}
    \nabla_{x} \log p(x|c) = \gamma(\nabla_x \log p(x|c) - \nabla_x \log p(x)) + \nabla_x \log p(x),
\label{cfg-tradition}
\end{equation}
where $c$ and $\gamma$ denote the condition and CFG coefficient, respectively. Unlike the condition of image understanding and generation models, which consists only of the text modality, the condition $c$ and the output $x$ in the interleaved multimodal generation scenario contains a sequence of texts and images, defined as:
\begin{equation}
   c = \{c_{txt}^0, c_{img}^0, c_{txt}^1, c_{img}^1, \dots\}, \quad x = \{x_{txt}^0, x_{img}^0, x_{txt}^1, x_{img}^1, \dots\}.
\end{equation}
Previous models~\cite{bagel} leverage classifier-free guidance
with respect to both text and image conditions, and randomly drop each $c_{txt}^j$ or $c_{img}^j$ independently during training. 
To address this complexity, we decompose the conditions into $c_{txt}=\{c_{txt}^j\}$, $c_{img}=\{c_{img}^j\}$ and the targets into $x_{txt}=\{x_{txt}^j\}$, $x_{img}=\{x_{img}^j\}$ according to their modalities, and then use an interleaved classifier-free guidance mechanism on image generation~\cite{mogao}, which is formulated as: 
\begin{equation}
\begin{aligned}
    \nabla_{x_{img}} \log p(x_{img}|c_{txt}, c_{img}) &= \gamma_{txt}(\nabla_{x_{img}} \log p(x_{img}|c_{txt}, c_{img})  \\ & - \nabla_{x_{img}} \log p(x_{img}|c_{img})) \\ & + \gamma_{img}(\nabla_{x_{img}} \log p(x_{img}|c_{img}) \\ & - \nabla_{x_{img}} \log p(x_{img})) + \nabla_{x_{img}} \log p(x_{img}).
\end{aligned}
\label{cfg-interleaved}
\end{equation}

In this way, we simplify the two-stage guidance formulation, with text-condition CFG as the core anchor and separate scaling coefficients $\gamma_{txt}, \gamma_{img}$ for image conditions. 
During training, we randomly take three condition sampling strategies on the conditions: keep both text and image conditions, only drop the text condition, and drop both text and image conditions simultaneously. 
Crucially, to maintain the model's ability to process the special tokens in the context, we exclude losses for text tokens and special tokens when we drop any conditions. As such, we can adjust the text guidance scale $\gamma_{txt}$ and image guidance scale $\gamma_{img}$ in inference to enhance image generation quality.

\subsection{Training Strategy}
\label{training_strategy}
Pre-trained unified MLLMs are typically optimized across multimodal understanding and generation tasks. To adapt pretrained unified MLLMs for diverse interleaved text-image tasks, we use a supervised training pipeline with two complementary data components under a unified training framework.

\textbf{In-context Generation Training.} 
This data component centers on in-context image generation samples.
It enhances the model's generative capabilities, which in turn alleviates the optimization challenges typically encountered during interleaved multimodal training. Moreover, it effectively addresses the performance stagnation when generating multiple sequential images by ensuring the model maintains dynamic generation quality throughout the process.

\textbf{Interleaved Text-Image Training.} 
We further incorporate a mixed corpus for interleaved text-image joint training, encompassing data of image understanding, image generation, and various interleaved text-image tasks. 
This hybrid training approach enables the model to achieve a comprehensive balance across different objectives. By training on these heterogeneous datasets simultaneously, the model can effectively strengthen its cross-task generalization and improve its performance across a wide range of tasks.

\section{Interleaved Data Curation}
\label{Data_curation}
For in-context generation training, we use datasets established in previous research~\cite{omnigen2,bagel}. Regarding interleaved text-image training, we aggregate existing data from VINCIE~\cite{vincie}, SEED-Story~\cite{storystream}, and CoMM~\cite{chen2025comm}. However, these open-source resources frequently exhibit limitations such as mediocre image quality, constrained instruction fidelity, and lack of task diversity. To address these shortcomings and align with our research objectives, we have curated an additional new comprehensive training dataset. The following sections elaborate on our data construction via three methods.

\begin{figure}[t]
    \centering
    \includegraphics[width=\textwidth]{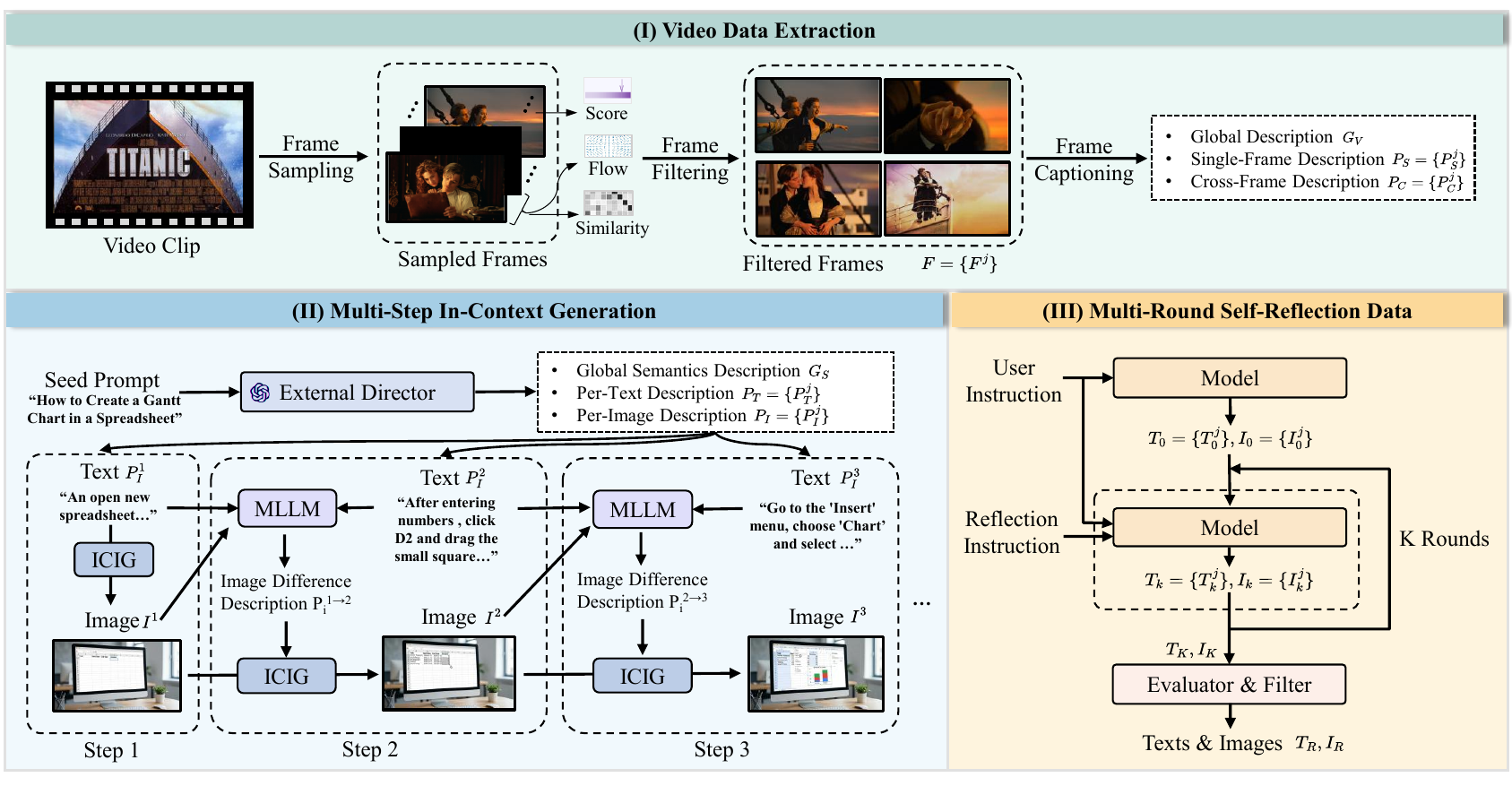}
    \caption{\textbf{The overall architecture of our data pipeline.} ``MLLM'' and ``ICIG'' represent multimodal large language model and in-context image generator, respectively, used for generating textual descriptions and images.} 
    \label{fig:datapipe}
\end{figure}

\subsection{Video Data Extraction}
Videos inherently encompass diverse visual information, including variations in character actions, object states, and scene viewpoints. Effective construction of high-quality interleaved text and image data requires the extraction of representative frames that capture major semantic transitions in chronological order and providing their associated textual descriptions rich in causal logic. Our approach addresses this need by developing a multi-stage video data extraction pipeline comprising three main procedures: frame sampling, frame filtering, and frame captioning. Figure~\ref{fig:datapipe} (I) illustrates this procedure to extract data from videos.

\textbf{Frame Sampling.}
Common frame sampling strategies, such as key-frame and uniform sampling \cite{omnigen2}, often struggle to align with actual visual content. Specifically, key-frames may not accurately reflect semantic shifts, resulting in uneven sample distributions or inconsistent semantic gaps. Furthermore, standard uniform sampling struggles with varying content velocities across different video segments, which may lead to significant information loss in dynamic sections. Considering these limitations, we propose a hybrid approach that implement a multi-interval sliding window sampling strategy to capture visual dynamics at various temporal scales to ensure both representativeness and effectiveness of the sampled frames. In practice, we uniformly sample adjacent k-frame windows (e.g., 2, 5 or 8 frames) at varying intervals (e.g., every 1, 3, or 10 seconds), and subsequently refine this selection through a multi-perspective filtering process. This strategy effectively captures the most informative frames while maintaining a coherent temporal flow. 

\textbf{Frame Filtering.}
To ensure high visual quality and minimize redundancy between adjacent frames, we implement a dual-stage content-based filtering process focusing on aesthetic quality and motion dynamics. 
(i) Aesthetic Filtering: To prioritize visual quality, we calculate each frame's Laplacian variance and the MANIQA~\cite{maniqa} metric as aesthetic score. Frames falling below a predefined score threshold are discarded as low-quality samples. 
To align with human preferences and ensure semantic coherence, we further utilize Qwen-3-VL-32B \cite{qwen3vl} as a high-level evaluator to determine whether each frame should be discarded or preserved based on its content integrity.
(ii) Motion Filtering: To eliminate the redundant frames, we use RAFT~\cite{raft} to estimate the optical flow between the source frame and target frame from given adjacent frames. Based on the computed flow fields, we exclude the target frames exhibiting either low mean or low variance, which typically indicates static content or signifies uniform global motion (e.g., camera translation or jitter) rather than meaningful object-level dynamics. To further reduce redundancy, we utilize DINOv2~\cite{divon2} to extract robust feature representations from the adjacent frame pairs and evaluate their feature differences, thus frames with negligible differences are filtered out as redundant. In this way, we obtain multiple images $F=\{F^j\}$ for one piece of data.

\textbf{Frame Captioning.}
To achieve comprehensive frame captioning, we propose a multi-level descriptive method that integrates intra-frame description with inter-frame analysis. Specifically, we leverage the MLLM Qwen-3-VL-32B \cite{qwen3vl} to generate descriptions across three granularities:
(i) Global Descriptions: These aim to synthesize the overarching narrative and thematic blueprint from the sequence of consecutive frames, which goes beyond literal visual mapping, instead emphasizing causal reasoning, temporal coherence, and the logical completeness of the depicted actions.
(ii) Single-frame Descriptions: These capture static visual attributes, including meticulous details such as color, quantity, morphology, and the spatial arrangement of objects relative to their background.
(iii) Cross-frame Descriptions: These focus on intricate inter-frame dynamics, encompassing: (1) subject-level dynamics, such as shifts in posture, orientation, facial expressions, and motion magnitude; (2) interaction states, involving variations in tools, targets, or agents; and (3) environmental nuances, including subtle transitions in lighting, shadows, and spatial structure.

In this way, we collect the frames$\{F^j\}$ and descriptions $G_V$, $P_S$, $P_C$ as a term of interleaved text-image data, serving to train the model’s capacity to learn the temporal evolution of actions or events.

\subsection{Multi-Step In-Context Generation}
The advanced image generation model has strong in-context generation abilities. It is able to extract visual concepts, such as objects or identities, from images and reproduce them in new images. This ability allows for the automatic creation of multi-image sequences with internal associations, serving as interleaved image-text data.
To construct this dataset, we develop a multi-step collaborative workflow. This workflow integrates an external MLLM as a director to implement chain-of-thought (CoT) reasoning and imagination for analytical understanding, and adopts an in-context image generator to achieve high-quality final image synthesis.

Figure~\ref{fig:datapipe} (II) illustrates the multi-step generation pipeline involves a CoT-based textual planning phase followed by iterative synthesis. First, an external MLLM (\eg, GPT-5~\cite{singh2025openai-gpt5}) creates a textual blueprint includes the global semantics description $G_S$, per-text description $P_T=\{P_T^j\}$ and per-image description $P_I=\{P_I^j\}$. 
Next, the image sequence is generated iteratively based on these descriptions. The first image $I^1$ is generated directly from its text prompt $P_I^1$.
After that, subsequent images are produced through a feedback loop. Specifically, we employ Qwen3-VL-32B\cite{qwen3vl} as another MLLM to analyze and generate image-to-image semantic differences that describe the visual changes needed from one image $I^j$ to the next $I^{j+1}$. The image difference description, along with the previous visual context $I^j$, are fed into the in-context image generator (\eg, Gemini 3 Pro~\cite{comanici2025gemini2.5}) to guide the precise image synthesis, resulting in the next image $I^{j+1}$.

Finally, this process creates interleaved data that includes global description $G_S$, texts $P_T=\{P_T^j\}$, images $I=\{I^j\}$, and texts $P_I=\{P_I^j\}$. 
Trained on such curated data, models are able to capture the chain-of-thought logic underlying the evolution of text-image narratives and corresponding visual variations.

\subsection{Multi-Round Self-Reflection Data} 
Inspired by recent advances in test-time scaling~\cite{lightman2024lets} and self-reflection for unified multimodal models~\cite{omnigen2}, the paradigms are extended to  interleaved multimodal generation to explore how iterative refinement can build text and image sequence. 

To construct the self-reflection dataset, the process is initiated by generating initial texts $T_0$ and images $I_0$ from curated user instructions. 
A powerful external MLLM (e.g., Gemini 3 Pro~\cite{comanici2025gemini2.5}) then acts as a critic, evaluating these outputs against the original requirements. Upon detecting instruction misalignment or aesthetic deficiencies, the critic diagnoses specific flaws and formulates rectification guidance. This feedback is then prepended to the original user prompt to steer the model through rounds of iterative refinement, culminating in the final reflective data $T_K, I_K$. Finally, these data are further evaluated by the external multimodal model to discard the suboptimal samples, ensuring high data fidelity and yielding $T_R$ and $I_R$.

\section{Experiments}
In this section, we conduct extensive experiments to validate the effectiveness of ILLUME-X as a unified model across a diverse set of vision–language tasks. 
We primarily evaluate the model on interleaved text-image generation and further benchmark its performance on standard text-to-image generation tasks.

\begin{figure}[h]
    \centering
    \includegraphics[width=\linewidth]{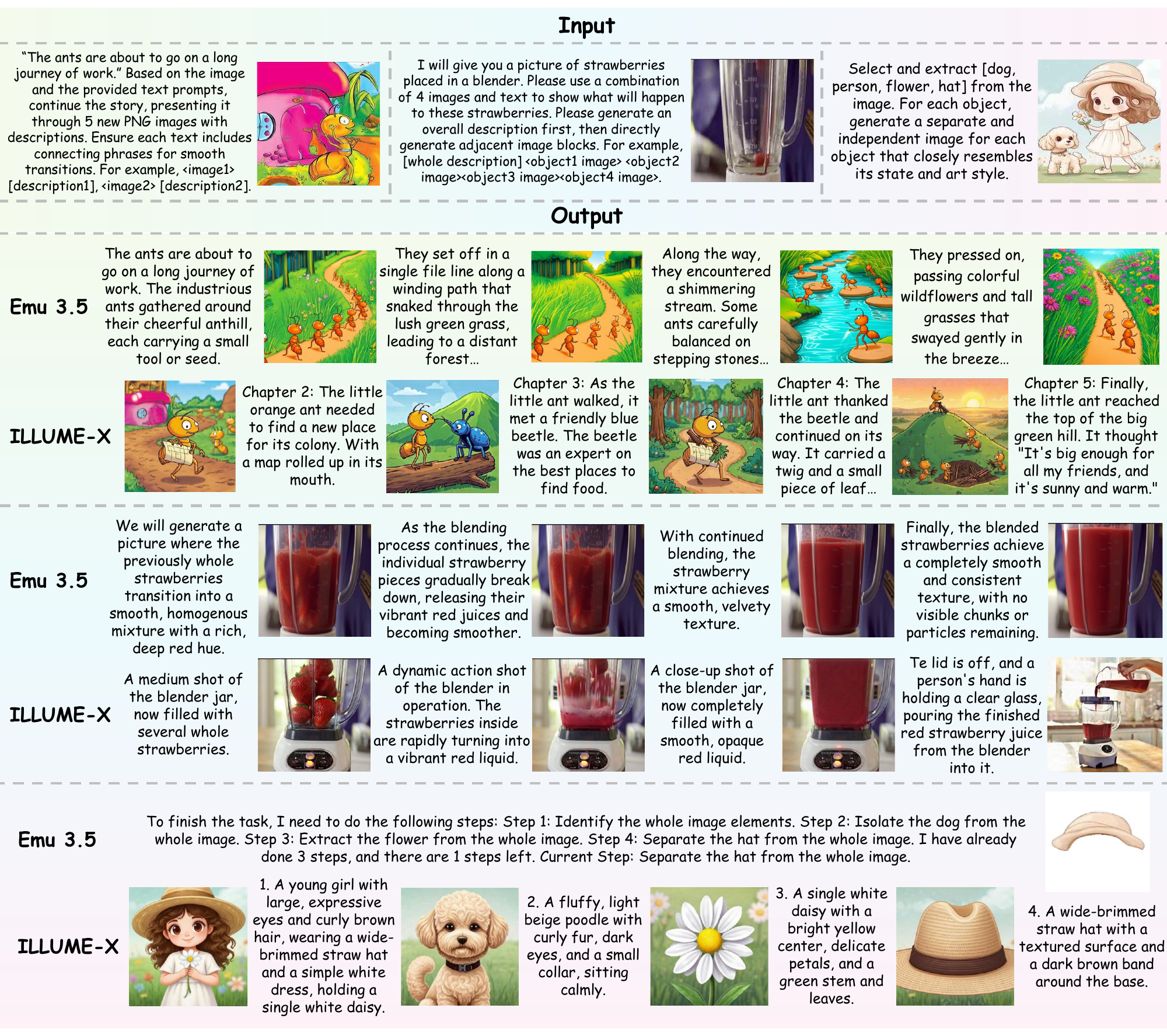}
    \caption{\textbf{Qualitative comparison with other methods.} We only show the first 4 groups when more than 4 generations are available.}
    \label{fig:main}
\end{figure}

\subsection{Evaluation}
\label{evaluation-main-text}

The widely adopted benchmark for interleaved text-image generation, ISG-Bench~\cite{chen2024interleaved-isg}, has notable evaluation limitations. Several of its metrics heavily depend on the structure of model outputs. As a result, when the structure of the generated results deviates from the predefined template, the evaluation mechanisms tend to fail. This makes it challenging to obtain an objective and robust measurement of a model’s interleaved text-image generation capability. To address these issues, we systematically extend and refine its evaluation dimensions and propose a novel metric termed ILScore.

\textbf{ILScore.} 
We present a hierarchical quantitative evaluation metric for interleaved text-image generation, which assesses both the global and local quality of generated text-image content from four distinct dimensions:
1) \textbf{image-text accuracy}, which measures overall text-image alignment in terms of coherence, relevance, and logical consistency;
2) \textbf{single image accuracy}, which evaluates each image based on concept consistency, relationship matching, detail fidelity, visual clarity, and aesthetics;
3) \textbf{image sequence accuracy}, which assesses cross-image content and stylistic consistency within multi-image sequences; and
4) \textbf{text accuracy}, which independently evaluates the quality and expressiveness of text-only outputs.
Together, these four dimensions provide complementary and holistic coverage, enabling precise evaluation across overall text-image alignment, single-image quality, multi-image coherence, and text-only accuracy.

\subsection{Implementation details}
We choose BAGEL~\cite{bagel} as the initialization of our model weights for its superior performance and public availability. The model is trained with open datasets such as SEED-Story~\cite{storystream}, VINCIE~\cite{vincie}, WEAVE~\cite{chow2025weave}, and our 100K curated dataset. The base learning rate is set to $2\times 10^{-5}$.  

\begin{table}[h]
\centering
\caption{\textbf{Quantitive comparison across different tasks on ISG-Bench~\cite{chen2024interleaved-isg}.}}
\label{tab:isg}
\scriptsize
\resizebox{\linewidth}{!}{
    \begin{tabular}{lcccccccccc}
    \toprule
    Model & \makecell{Is unified \\ model?} & \makecell{Style \\ Transfer} & Progressive & 3D\_Scene & \makecell{Image \\ Decomposition} & \makecell{Image-Text \\ Complementation} & \makecell{Temporal\\Prediction} & \makecell{Visual\\StoryTelling} & VQA & \textbf{AVG} \\
    \hline
    Show-o~\cite{xie2024show-o} & \ding{51} & 2.11 & 2.41 & 1.43 & 2.87 & 2.06 & 2.58 & 3.32 & 1.86 & 2.33 \\
    Anole~\cite{chern2024anole} & \ding{51} & 2.93 & 2.76 & 1.85 & 1.49 & 3.21 & 2.58 & 2.97 & 4.70 & 2.81 \\
    Minigpt-5~\cite{zheng2023minigpt} & \ding{51} & 2.15 & 3.15 & 1.79 & 2.54 & 2.72 & 2.73 & 2.91 & 4.29 & 2.79 \\
    CoMM-Minigpt-5~\cite{chen2025comm}  & \ding{51} & 2.96 & 2.60 & 3.09 & 2.24 & 3.09 & 2.52 & 2.72 & 2.87 & 2.96 \\
    Seed-Llama-14b~\cite{ge2023making-seed-llama}  & \ding{51} & 1.84 & 3.30 & 1.52 & 3.69 & 1.94 & 1.78 & 2.84 & 2.20 & 2.39 \\
    Gemini~\cite{team2024gemini-01} \& SD3~\cite{esser2024scaling-sd3} & \ding{55} & 4.89 & \textbf{6.59} & 2.68 & 7.26 & 6.37 & 5.26 & 5.68 & \textbf{7.89} & 5.83 \\
    ISG-AGENT~\cite{chen2024interleaved-isg} & \ding{55} & 5.87 & 6.46 & \textbf{4.89} & \textbf{7.58} & 6.93 & 4.54 & 7.03 & 6.80 & 6.26 \\
    \hline
    \rowcolor{gray!20}
    ILLUME-X & \ding{51} & \textbf{6.02} & 5.75 & 3.96 & 6.97 & \textbf{7.06} & \textbf{6.43} & \textbf{7.25} & 6.67 & \textbf{6.26} \\
    \bottomrule
    \end{tabular}
    }
\end{table}
\begin{table}[t]
\centering
\caption{
\textbf{Quantitative evaluation of various models across the four ILScore dimensions.} We report performance across eight sub-tasks and calculate the average score for: D1 (image-text accuracy), D2 (single image accuracy), D3 (image sequence accuracy), and D4 (text accuracy). Bold values indicate the best overall performance in each category.
}
\label{tab:interleaved}
\scriptsize
\resizebox{\linewidth}{!}{
    \begin{tabular}{>{\centering\arraybackslash}p{0.5cm}|lcccccccccc}
    \toprule
    & Model & \makecell{Style \\ Transfer} & Progressive & \makecell{3D \\ Scene} & \makecell{Image \\ Decomposition} & \makecell{Image-Text \\ Complementation} & \makecell{Temporal\\Prediction} & \makecell{Visual\\StoryTelling} & VQA & \textbf{AVG} \\
    \midrule
    
    \multirow{3}{*}{\rotatebox[origin=c]{90}{D1}}
    & Emu 3.5~\cite{emu35}  & 2.16  &  4.20 & 2.50 &  2.67  &  6.40  &  5.40  & 4.42  & 5.00  & 4.09 \\
    & Gemini 3 pro~\cite{comanici2025gemini2.5} & 4.10 & 3.93 & 3.38 & 1.11 & 4.00 & 3.40 & 4.64 & 2.00 & 3.32  \\
    & ILLUME-X & 3.80 & 3.60 & 3.12 & 4.60 & 4.56 & 3.80 & 6.64 & 2.40 & 4.07  \\
    \midrule
    
    \multirow{3}{*}{\rotatebox[origin=c]{90}{D2}}
    & Emu 3.5  & 7.44 & 7.19 & 4.80 & 5.48 & 7.58 & 7.73 & 7.27 & 5.75 & 6.66  \\
    & Gemini 3 pro  & 4.42 & 6.51 & 5.08 & 4.50 & 9.42 & 7.46 & 4.00 & 9.76 & 6.39  \\
    & ILLUME-X  & 6.33 & 5.84 & 5.93 & 7.21 & 6.41 & 5.96 & 7.44 & 6.41 & 6.44  \\
    \midrule
    
    \multirow{3}{*}{\rotatebox[origin=c]{90}{D3}}
    & Emu 3.5  & 5.05 & 3.93 & 3.50 & 4.70 & 7.00 & 8.20 & 4.25 & 2.30 & 4.87  \\
    & Gemini 3 pro  & 2.75 & 6.20 & 2.38 & 3.22 & 3.89 & 7.30 & 4.18 & 4.80 & 4.34 \\
    & ILLUME-X  & 6.25 & 4.47 & 6.12 & 4.60 & 5.89 & 6.00 & 5.93 & 5.70 & 5.62  \\
    \midrule
    
    \multirow{3}{*}{\rotatebox[origin=c]{90}{D4}}
    & Emu 3.5  & 2.70 & 5.57 & 5.25 & 2.40 & 8.80 & 8.00 & 5.58 & 7.20 & 5.69  \\
    & Gemini 3 pro  & 6.50 & 5.33 & 5.62 & 2.56 & 5.44 & 5.10 & 6.55 & 5.20 & 5.29 \\
    & ILLUME-X  & 4.25 & 4.53 & 4.62 & 5.60 & 6.33 & 5.90 & 7.93 & 2.80 & 5.25 \\
    \midrule

    \multirow{3}{*}{\rotatebox[origin=c]{90}{Overall}}
    & Emu 3.5 & 4.34 & 5.22 & 4.01 & 3.81 & \textbf{7.45} & \textbf{7.33} & 5.38 & 5.06 & 5.33 \\
    & Gemini 3 pro  & 4.44 & \textbf{5.49} & 4.12 & 2.85 & 5.69 & 5.82 & 4.84 & \textbf{5.44} & 4.84 \\
    
     & \cellcolor{gray!20}ILLUME-X & \cellcolor{gray!20}\textbf{5.16} & \cellcolor{gray!20}4.61 & \cellcolor{gray!20}\textbf{4.95} & \cellcolor{gray!20}\textbf{5.50} & \cellcolor{gray!20}5.80 & \cellcolor{gray!20}5.42 & \cellcolor{gray!20}\textbf{6.99} & \cellcolor{gray!20}4.33 & \cellcolor{gray!20}\textbf{5.34} \\
    
    \bottomrule
    \end{tabular}
}

\end{table}
\begin{table}[h]
    \centering
    \caption{\textbf{Cost Comparison.}}
    \resizebox{0.5\linewidth}{!}{
        \begin{tabular}{c|cc}
        \toprule
             Model & Parameters & Time per Image \\
        \midrule
             Emu 3.5 & 34B & 409.50s \\
             \rowcolor{gray!20} ILLUME-X (Ours) & 7B + 7B & 81.33s \\
        \bottomrule
        \end{tabular}
    }
    \label{tab:cost-compare}
\end{table}

\subsection{Main Results}

\subsubsection{Qualitative Comparison on Interleaved Text-Image Generation.} 
As shown in Figure~\ref{fig:main},
ILLUME-X significantly excels in narrative consistency and complex instruction following compared to Emu 3.5~\cite{emu35}. Specifically, in the first group, ILLUME-X generates a more precise and cohesive narrative with a visual representation that seamlessly corresponds to the description of the journey of the ants, showcasing its strength in visual storytelling. Similarly, in the second group, ILLUME-X produces a smooth transition from whole strawberries to blended juice, illustrating its strength in handling complex transformations in image generation. In contrast, Emu 3.5~\cite{emu35} often struggles with more abstract connections between text and image, resulting in less coherent outputs in some cases. Overall, ILLUME-X demonstrates superior performance in generating visually aligned and narratively consistent results across different tasks, making it a strong contender in interleaved multimodal generation tasks.


\subsubsection{Quantitative Comparison on Text-Image Generation.} 
For the interleaved text-image task, we first follow the popularly used ISG-Bench~\cite{chen2024interleaved-isg} to quantitatively evaluate interleaved text-image generation. As illustrated in Table~\ref{tab:isg}, our proposed ILLUME-X achieves state-of-the-art (SOTA) performance among all unified models, reaching an average score of 6.26. Notably, ILLUME-X not only outperforms existing unified models like Anole (2.81) and MiniGPT-5 (2.79) by a significant margin (over $120\%$ improvement in AVG score) but also matches the performance of the non-unified agent-based system, ISG-AGENT (6.26). This demonstrates the effectiveness of our approach in handling complex interleaved generation tasks within a single unified framework.

Then, we evaluate our ILLUME-X, open-source Emu 3.5 and commercial models Gemini 3 Pro on our proposed ILScore. As illustrated in Table~\ref{tab:interleaved}, ILLUME-X overall achieves the highest aggregate score (5.34), marginally surpassing Emu 3.5 (5.33) and significantly outperforming Gemini 3 Pro (4.84). ILLUME-X demonstrates superior robustness in complex structural tasks, securing top honors in four out of eight categories: Style Transfer, 3D Scene, Image Decomposition, and Visual Storytelling.
%
More notably, our method substantially reduces training overhead and achieves markedly lower inference latency, as validated in Table~\ref{tab:cost-compare}, making it better suited for practical real-world deployment scenarios.

\begin{table}[h]
\centering
\caption{\textbf{Comparison of different models on GenEval~\cite{ghosh2023geneval} benchmark.}}
\label{tab:GenEval}
\scriptsize
\resizebox{\linewidth}{!}{
    \begin{tabular}{lcccccccc}
    \toprule
    Model & \# Params & \multicolumn{7}{c}{GenEval~\cite{ghosh2023geneval}} \\
    & B & Single object & Two objects & Counting & Colors & Position & Color attribution & Overall \\
    \hline
    SDXL~\cite{podell2024sdxl} & 6.6 & 0.98 & 0.74 & 0.39 & 0.85 & 0.15 & 0.23 & 0.55 \\
    FLUX.1-dev~\cite{labs2025flux1kontextflowmatching} & 8+12 & 0.99 & 0.81 & 0.79 & 0.74 & 0.20 & 0.47 & 0.67 \\
    \hline
    Show-o~\cite{xie2024show-o}  & 1.3 & 0.98 & 0.80 &0.66 &0.84 &0.31 &0.50 &0.68 \\
    Janus-Pro~\cite{januspro} & 7 & 0.99 & 0.89 & 0.59 & 0.90 & \textbf{0.79} & 0.66 & 0.80 \\
    UniWorld-V1~\cite{lin2025uniworld} & 7+12 & 0.99 & 0.93 & 0.79 & 0.89 & 0.49 & 0.70 &0.80 \\
    OmniGen2~\cite{omnigen2} & 3+4 & \textbf{1.00} & \textbf{0.95} & 0.64 & 0.88 & 0.55 & \textbf{0.76} & 0.80 \\
    BAGEL~\cite{bagel} & 7+7 & 0.99 & 0.94 & 0.81 & 0.88 & 0.64 & 0.63 & 0.82 \\
    \hline
    \rowcolor{gray!20}
    ILLUME-X & 7+7 & 0.99 & \textbf{0.95} & \textbf{0.81} & \textbf{0.92} & 0.71 & 0.74 & \textbf{0.85} \\
    \bottomrule
    \end{tabular}
    }
\end{table}
\begin{table}[h]
\centering
\caption{\textbf{Comparison of different models on DPG-Bench~\cite{hu2024ella-dpg-bench} benchmark.}}
\label{tab:DPG}
\scriptsize
\resizebox{\linewidth}{!}{
    \begin{tabular}{lccccccc}
    \toprule
    Model & \# Params & \multicolumn{6}{c}{DPG-Bench~\cite{hu2024ella-dpg-bench}}  \\
    & B & \ \ Global\ \ & \ \ Entity\ \  & \ \ Attribute\ \  & \ \ Relation\ \  & \ \ Other\ \  & \ \ Overall\ \  \\
    \hline
    SDXL~\cite{podell2024sdxl} & 6.6 & 83.27 & 82.43 & 80.91 & 86.76 & 80.41 & 74.65 \\
    FLUX.1-dev~\cite{labs2025flux1kontextflowmatching} & 8+12 & 82.1& 89.5 &88.7& \textbf{91.1}& 89.4 &84.0 \\
    \hline
    Show-o~\cite{xie2024show-o}  & 1.3 & 79.33 &75.44 &78.02 &84.45& 60.80& 67.27 \\
    Janus-Pro~\cite{januspro} & 7 & 86.90& 88.90 &89.40 &89.32& 89.48& 84.19 \\
    UniWorld-V1~\cite{lin2025uniworld} & 7+12 & 83.64 &88.39& 88.44& 89.27& 87.22& 81.38 \\
    OmniGen2~\cite{omnigen2} & 3+4 & 88.81 &88.83 &90.18 &89.37 &\textbf{90.27}& 83.57 \\
    BAGEL~\cite{bagel} & 7+7 &  88.94 &90.37& 91.29 &90.82 &88.67 &85.07 \\
    \hline
    \rowcolor{gray!20}
    ILLUME-X & 7+7 & \textbf{90.28} & \textbf{91.80} & \textbf{94.08} & 82.37 & 88.00 & \textbf{86.38} \\
    \bottomrule
    \end{tabular}
    }
\end{table}

\textbf{More Comparison on the Text-to-Image Task.}
We also assess the generation quality for the text-to-image task using widely adopted benchmarks, GenEval~\cite{ghosh2023geneval} and DPG-Bench~\cite{hu2024ella-dpg-bench}, to evaluate both basic semantic alignment and complex scene synthesis. 
As shown in Table~\ref{tab:GenEval}, ILLUME-X outperforms other models on the GenEval benchmark, achieving an overall score of 0.85, surpassing OmniGen2 (0.80) and BAGEL (0.82). ILLUME-X excels in Counting (0.81), Colors (0.92), and Overall performance (0.85), while also showing strong results in Single Object (0.99) and Two Objects (0.95). Its performance in Color Attribution (0.74) and Position (0.71) further underscores its effectiveness in handling complex scene synthesis, making it a competitive choice in text-to-image generation.

As shown in Table~\ref{tab:DPG}, it presents a comprehensive evaluation of image generation performance on DPG-Bench. ILLUME-X achieves the highest Overall score of 86.38, surpassing strong competitors such as BAGEL (85.07) and Janus-Pro (84.19). A key driver of this performance is our model’s exceptional accuracy in Attribute (94.08) and Entity (91.80) categories, where it outperforms all baseline models, including the significantly larger FLUX.1-dev. The substantial margin in the Attribute score (+2.79 over BAGEL) highlights ILLUME-X’s precision in grounding specific textual descriptors to visual entities.

Interestingly, while our model shows a slight performance gap in the Relation sub-category compared to FLUX.1-dev, its Global (90.28) score indicates a superior ability to maintain holistic scene coherence. Combined with its efficient 7B+7B parameter architecture, these results further validate the effectiveness of ILLUME-X in handling dense and complex prompts that require fine-grained semantic alignment.

\begin{table}[h]
\centering
\caption{\textbf{Ablation Results.} Quantitative Comparison of different CFG parameters and CoT settings for text-image interleaved generation. $\gamma_{txt}$ represents the text guidance scale, and $\gamma_{img}$ denotes the image guidance scale.}

\label{tab:ablation-table}
\scriptsize
\resizebox{0.8\linewidth}{!}{
    \begin{tabular}{ccc|cc}
    \toprule
    \quad $\gamma_{txt}$ \quad & $\gamma_{img}$ \quad & w/ CoT & Visual Storytelling & Image-Text Complementation \\
    \hline
     2.0 & 1.0 & \ding{55}&  7.02 & 5.35 \\
     4.0 & 1.0 & \ding{55} &  7.13 & 5.59 \\
     8.0 & 1.0 & \ding{55} &  7.12 & 5.69 \\ 
     8.0 & 1.25 & \ding{55} &  7.00 & 5.52 \\
     8.0 & 1.5 & \ding{55} &  6.85 & 5.60 \\

    \rowcolor{gray!20}
     8.0 & 1.0 & \ding{51} &  \textbf{7.28} & \textbf{5.91} \\
    \bottomrule
    \end{tabular}
}
\end{table}
\begin{figure}[h]
    \centering
    \includegraphics[width=\linewidth]{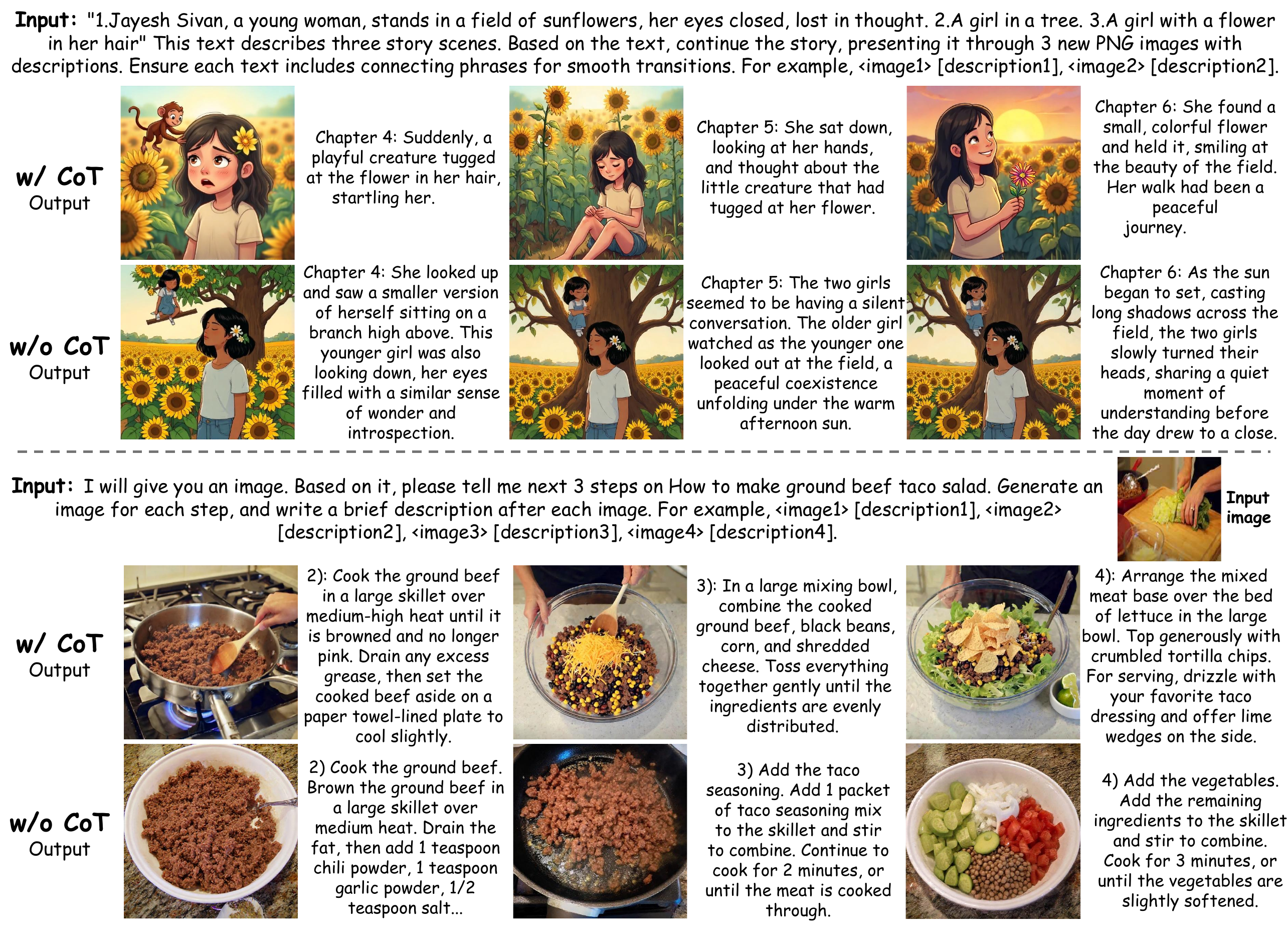}
    \caption{\textbf{Qualitative comparison of different CoT settings.}}
    \label{fig:Ablation-fig}
\end{figure}

\subsection{Ablation Study}
To efficiently and targetedly validate the effect of the interleaved text-image  structure, we conduct ablation studies only on the Visual Storytelling and ``howto'' subset of Image-Text Complementation. These two task categories inherently rely on temporal logic and strong multimodal complementarity, making them representative scenarios where interleaving is expected to yield the most pronounced gains. All remaining tasks adopt the best-performing structure identified through ablation for fair comparison. 
Experimental results demonstrate that the best performance is achieved when the CFG uses a text guidance scale of 8.0, an image guidance scale of 1.0, and incorporates CoT. We visualize qualitative examples with and without CoT under these optimal hyperparameters in Figure~\ref{fig:Ablation-fig}.

\section{Limitations}
ILLUME-X is primarily trained and evaluated for interleaved text-image generation at a resolution of 512. Due to limitations imposed by the underlying model architecture and the finite context length, scaling both training and inference to resolutions of 1024 and above remains challenging. Consequently, the quality of high-resolution interleaved generation still leaves room for further improvement.

\section{Conclusion}
In this work, we present ILLUME-X, a unified multimodal model that supports free-form interleaved text-image generation. We also build an efficient multimodal interleaving data construction pipeline to produce training samples. In addition, we extend existing evaluation metrics and introduce ILScore to better reflect the requirements of interleaved text-image generation. Extensive comparative experiments demonstrate the effectiveness of the proposed framework. 

\section{Acknowledge}
This work was supported by the National Cyber Security-National Science and Technology Major Project under Grant No. 2026ZD1500500 and the National Natural Science Foundation of China (NSFC) under Grant No.62476067.

%



%
%
\bibliographystyle{splncs04}
\bibliography{main}
\clearpage
\begin{appendix}
\setcounter{page}{1}
\hypersetup{hypertexnames=false}

\setcounter{figure}{0}  
\setcounter{table}{0}   
\renewcommand{\thefigure}{S\arabic{figure}} 
\renewcommand{\thetable}{S\arabic{table}}

\section*{Supplementary Material}

In this supplementary material, we provide additional explanation and experimental results to further support the main paper. The content is organized as follows: 
Sec.~\ref{sec:appendix_a} provides a more detailed experimental setup; 
Sec.~\ref{eval-detail} presents detailed prompt templates used in our evaluation framework; 
Sec.~\ref{more-results} offers additional experimental result visualizations;  
Sec.~\ref{our-datasets} provides some cases of custom interleaved datasets.

\section{Experiment Setting}
\label{sec:appendix_a}
Table~\ref{tab:training_recipe} introduce a more detail experimental setup, including hyperparameters and training data.
\begin{table}[htbp]
  \centering
  \small
  \caption{Training recipe of ILLUME-X.}
  \label{tab:training_recipe}
  \begin{tabular}{lc} 
    \toprule
    \textbf{Hyper Parameters} & \textbf{Parameter}\\
    Training Steps & 6005 \\
    Warmup Steps & 100 \\
    Resolution & 512 \\
    \hline
    \textbf{Data Setting} & \textbf{Num} \\
    Understanding & 9,902,211 \\
    Image Generation & 67,101,538 \\
    Interleaved Tasks & 446,482 \\
    \bottomrule
  \end{tabular}
\end{table}

\section{Evaluation}
\label{eval-detail}
\begin{figure}
    \centering
    \includegraphics[width=\linewidth]{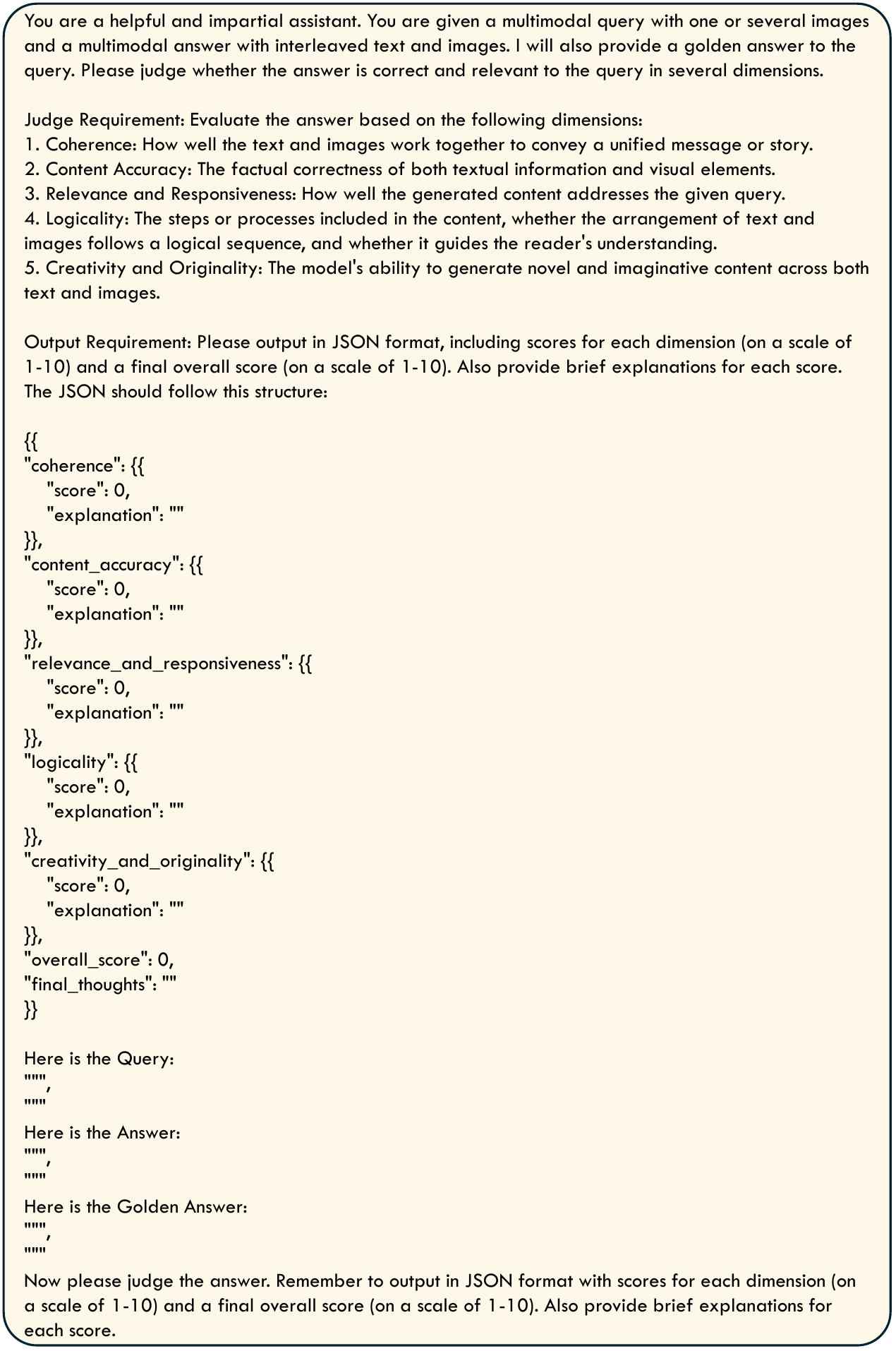}
    \caption{\textbf{Prompt of Image-Text Accuracy (w GT)}}
    \label{fig:image-text-gt}
\end{figure}

\begin{figure}
    \centering
    \includegraphics[width=\linewidth]{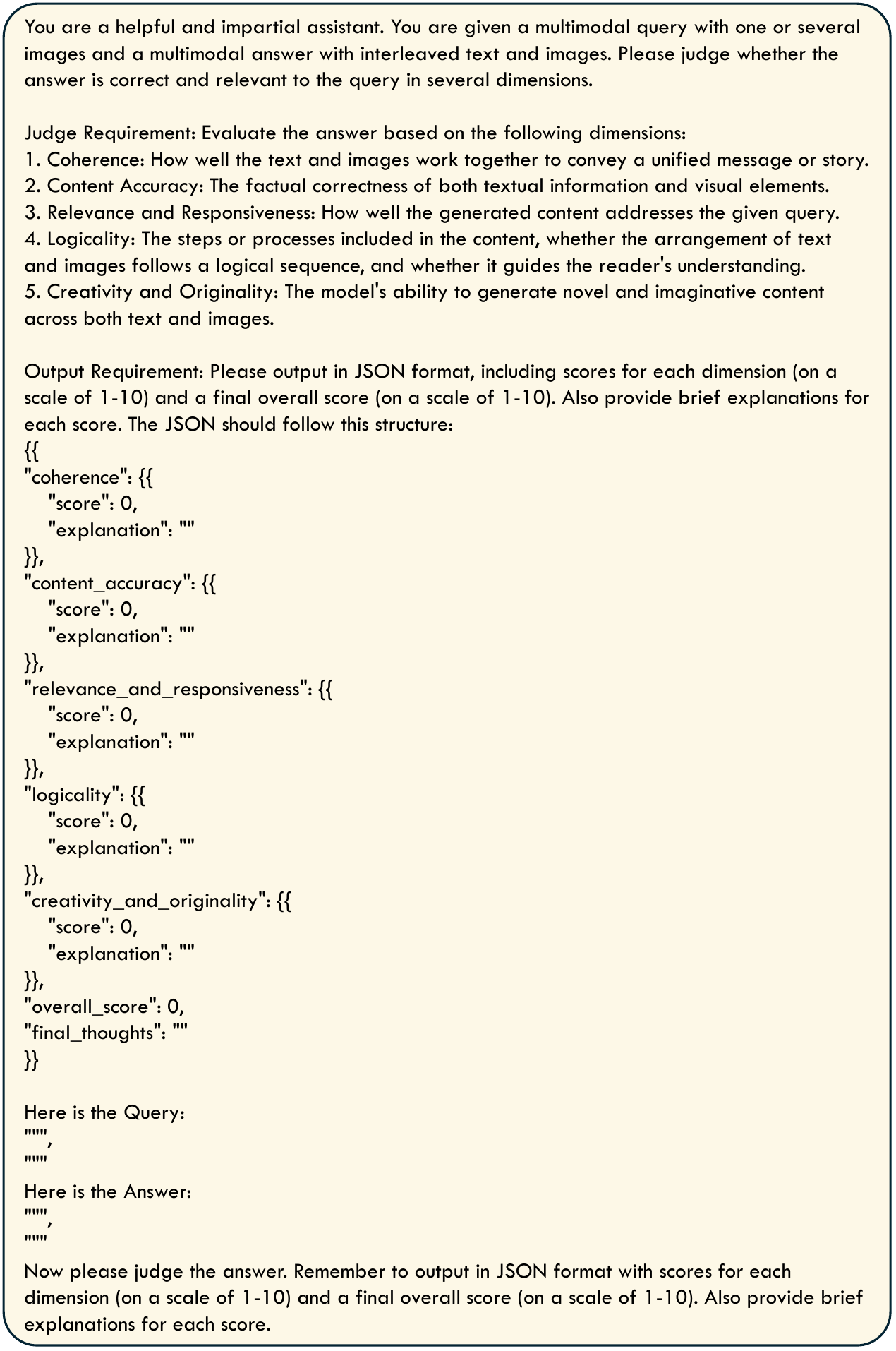}
    \caption{\textbf{Prompt of Image-Text Accuracy (w/o GT)}}
    \label{fig:image-text-wo-gt}
\end{figure}

\begin{figure}
    \centering
    \includegraphics[width=\linewidth]{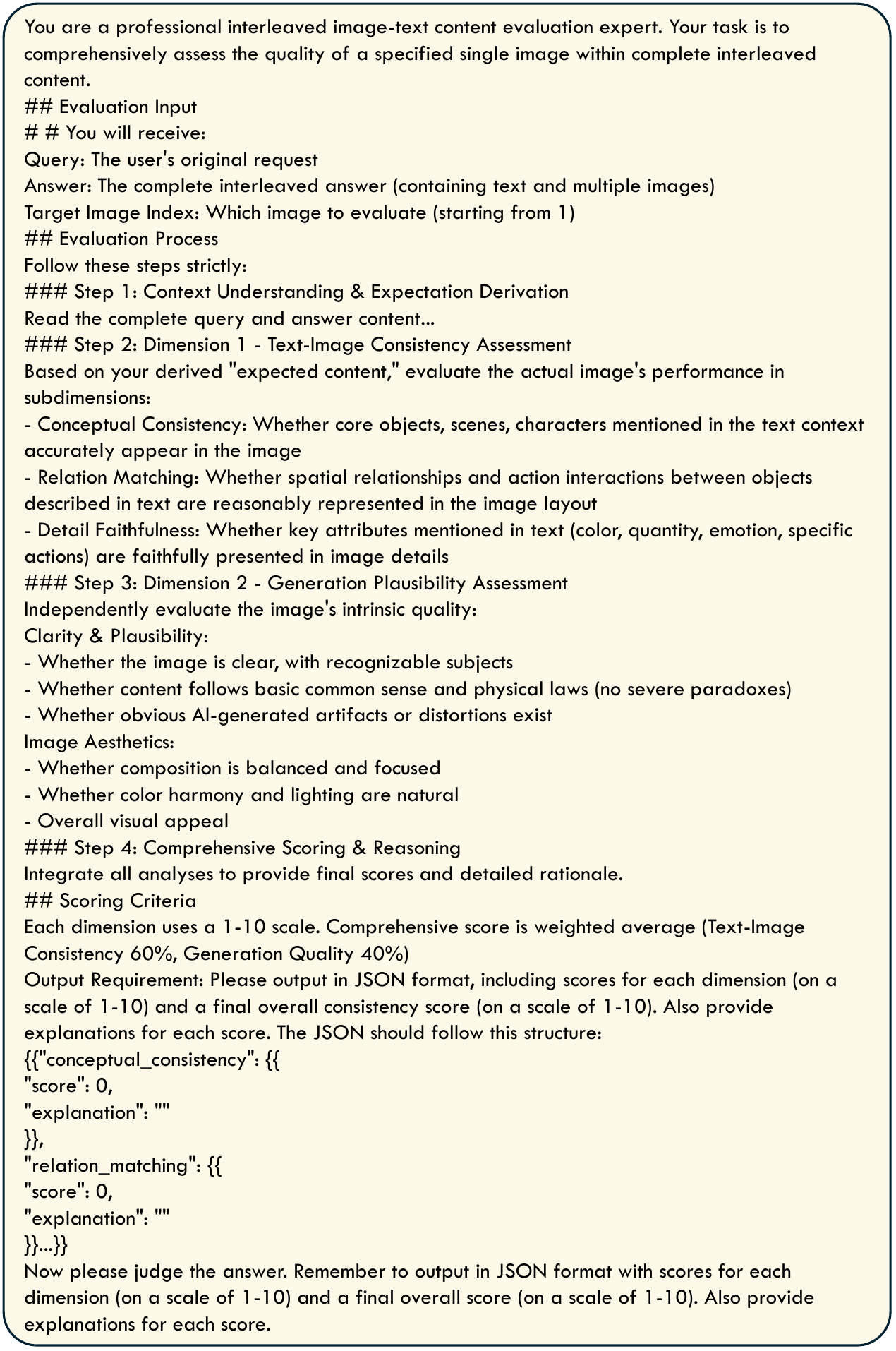}
    \caption{\textbf{Prompt of Single Image Accuracy}}
    \label{fig:single-image}
\end{figure}

\begin{figure}
    \centering
    \includegraphics[width=\linewidth]{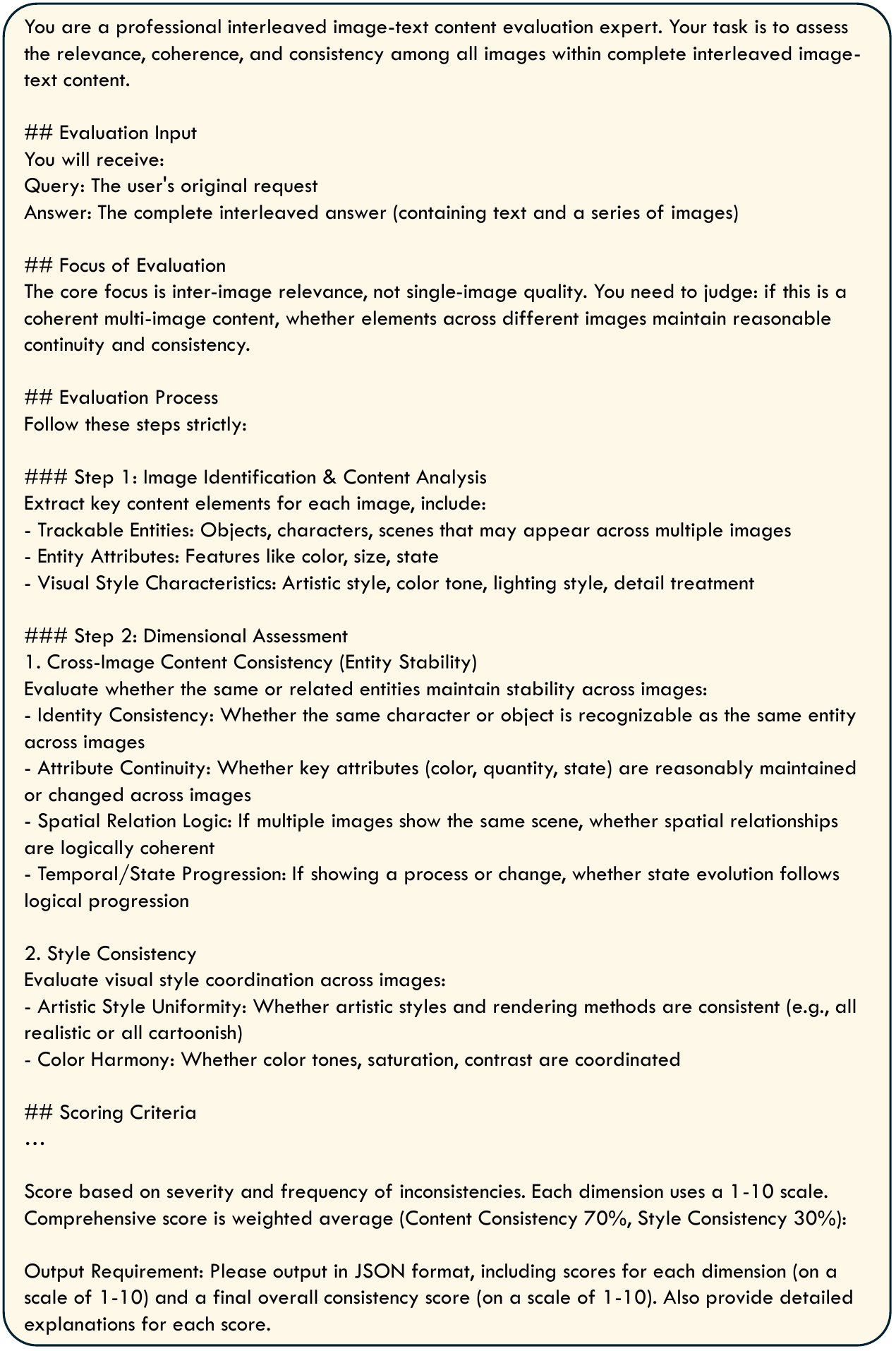}
    \caption{\textbf{Prompt of Image Sequence Accuracy}}
    \label{fig:image-sequence}
\end{figure}

\begin{figure}
    \centering
    \includegraphics[width=\linewidth]{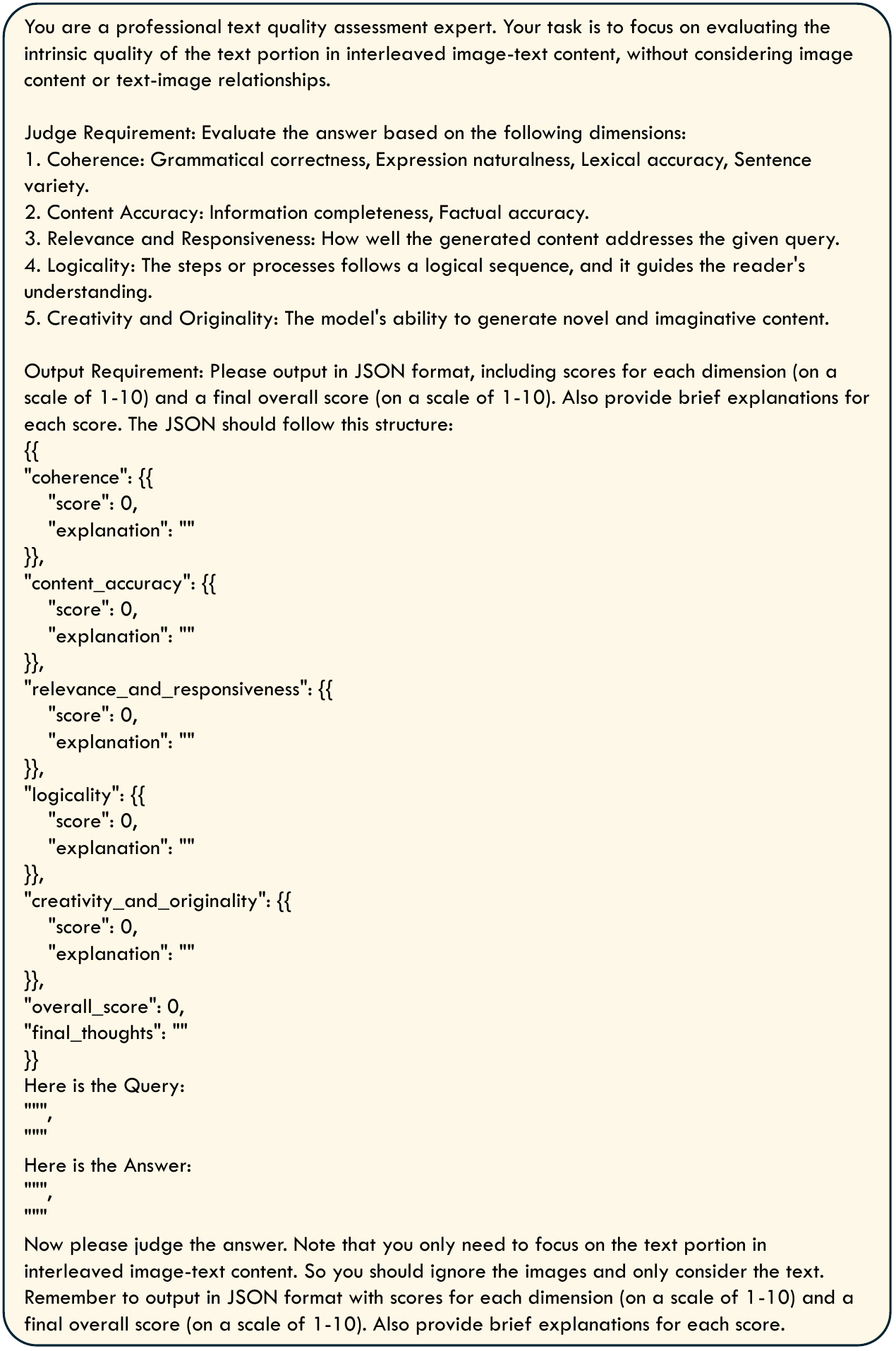}
    \caption{\textbf{Prompt of Text Accuracy}}
    \label{fig:txt-only}
\end{figure}

In Sec.~\ref{evaluation-main-text}, we introduce a new evaluation metric, ILScore, which comprehensively considers four accuracy dimensions, including image-text, single image, image sequence and text. Here, we provide a detailed introduction to ILScore.
\begin{itemize}
    \item \textbf{Image-Text Accuracy.} We define five evaluation metrics for multimodal generation quality: \textbf{\textit{coherence}}, \textbf{\textit{content accuracy}}, \textbf{\textit{relevance and responsiveness}}, \textbf{\textit{logicality}}, and \textbf{\textit{creativity and originality}}. Among them, \textbf{\textit{coherence}} measures whether the text and image are smoothly connected in theme and content, and whether they effectively integrate to convey a unified message or narrative; \textbf{\textit{content accuracy}} characterizes the factual correctness of textual information and visual elements; \textbf{\textit{relevance and responsiveness}} characterize the degree to which the generated content responds to the given query; \textit{Logicality} assesses whether the procedural steps, content flow, and layout of text and figures adhere to a rational and consistent order, thereby enabling readers to follow and comprehend the presented information effectively; \textit{Creativity and originality} assess whether the generated content is novel, distinctive, and sufficiently engaging to attract readers.
    
    \item \textbf{Single Image Accuracy.} This evaluation dimension mainly considers five aspects, including \textbf{\textit{conceptual consistency}}, \textbf{\textit{relation matching}}, \textbf{\textit{detail faithfulness}}, \textbf{\textit{clarity and plausibility}} and \textbf{\textit{image aesthetics}}. Here, \textbf{\textit{conceptual consistency}} evaluates whether the core objects, scenes, and characters described in the text appear accurately in the image; \textbf{\textit{relation matching}} measures whether the spatial relationships and action interactions among objects described in the text are reasonably reflected in the image layout; \textbf{\textit{detail faithfulness}} represents whether the key attributes mentioned in the text (\eg, color, quantity, emotion, specific actions) are faithfully presented in the image details; \textbf{\textit{clarity and plausibility}} focuses on whether the image is clear with identifiable subjects and the content conforms to common sense without obvious physical paradoxes; \textbf{\textit{image aesthetics}} measures whether composition, color, lighting, etc., possess basic aesthetic quality without severe distortions.
    
    \item \textbf{Image Sequence Accuracy.} This evaluation dimension mainly focuses on \textbf{\textit{cross image content consistency}} and \textbf{\textit{style consistency}}. \textbf{\textit{cross image content consistency}} assesses whether the appearance and identity of objects, characters, and scenes remain stable across different images of the same sequence; \textbf{\textit{style consistency}} requires a consistent artistic style across images of the same sequence.
    
    \item \textbf{Text accuracy.} Text accuracy adopts the same metric as image-text accuracy, but focuses exclusively on the accuracy of the text itself.
    
\end{itemize}
The detailed prompts we use for scoring are shown in Fig.~\ref{fig:image-text-gt},~\ref{fig:image-text-wo-gt},~\ref{fig:single-image},~\ref{fig:image-sequence}, and~\ref{fig:txt-only}.

\section{More Results}
\label{more-results}
\begin{figure}
    \centering
    \includegraphics[width=\linewidth]{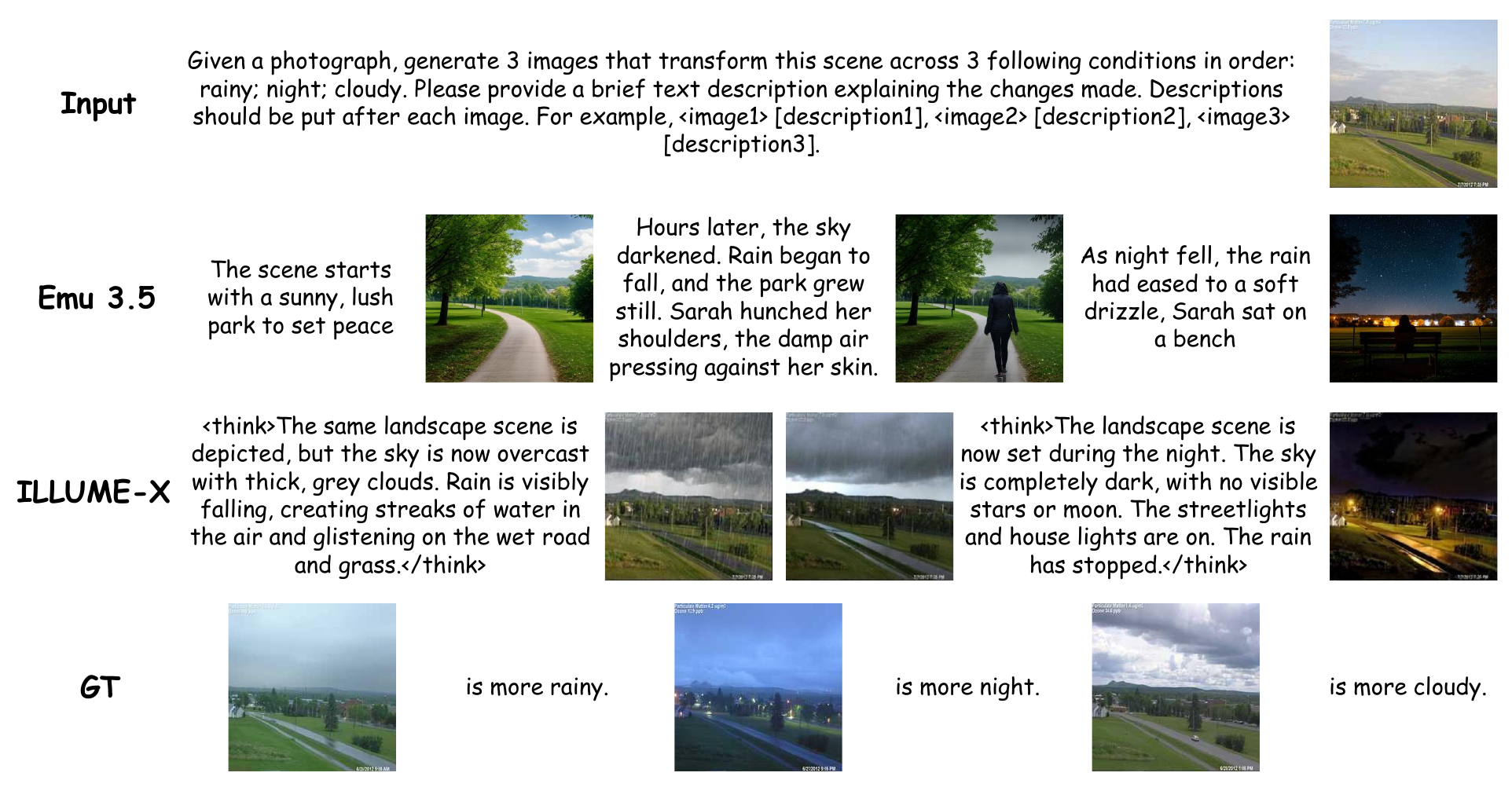}
    \caption{\textbf{Results of Style transfer}}
    \label{fig:style-transfer}
\end{figure}

\begin{figure}
    \centering
    \includegraphics[width=\linewidth]{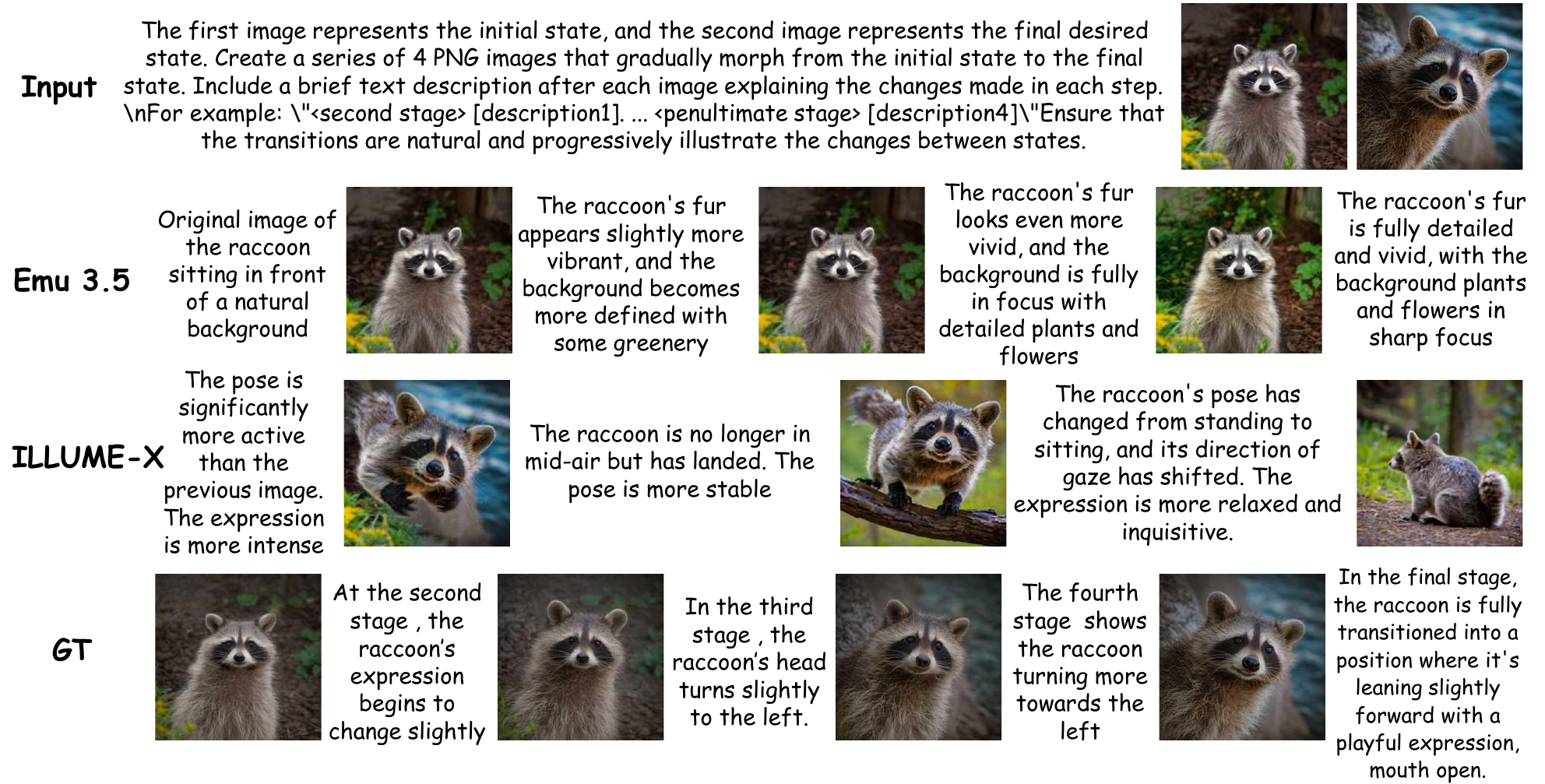}
    \caption{\textbf{Results of Progressive}}
    \label{fig:Progressive}
\end{figure}

\begin{figure}
    \centering
    \includegraphics[width=\linewidth]{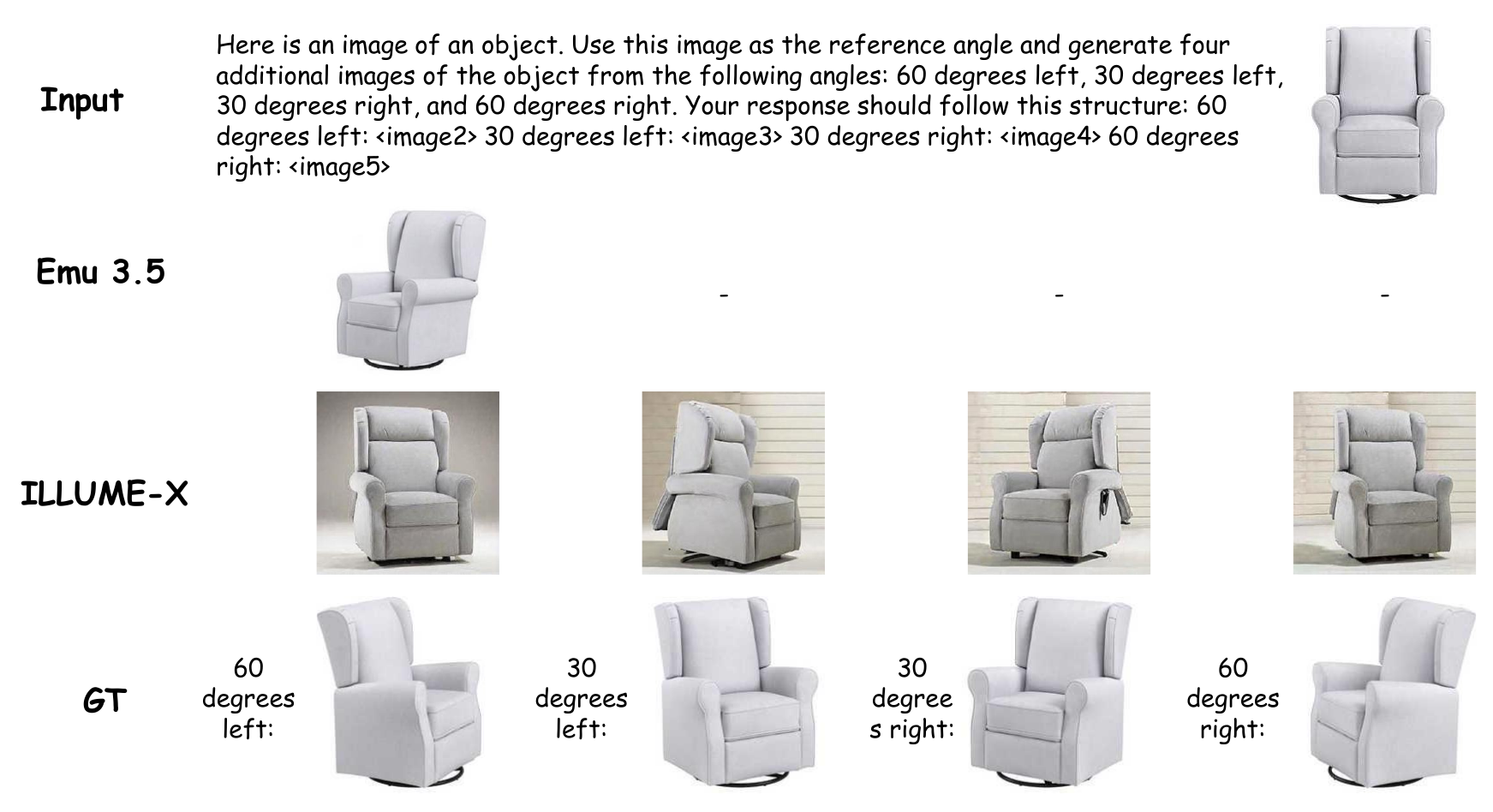}
    \caption{\textbf{Results of 3D Scene}}
    \label{fig:3dscene}
\end{figure}

\begin{figure}
    \centering
    \includegraphics[width=\linewidth]{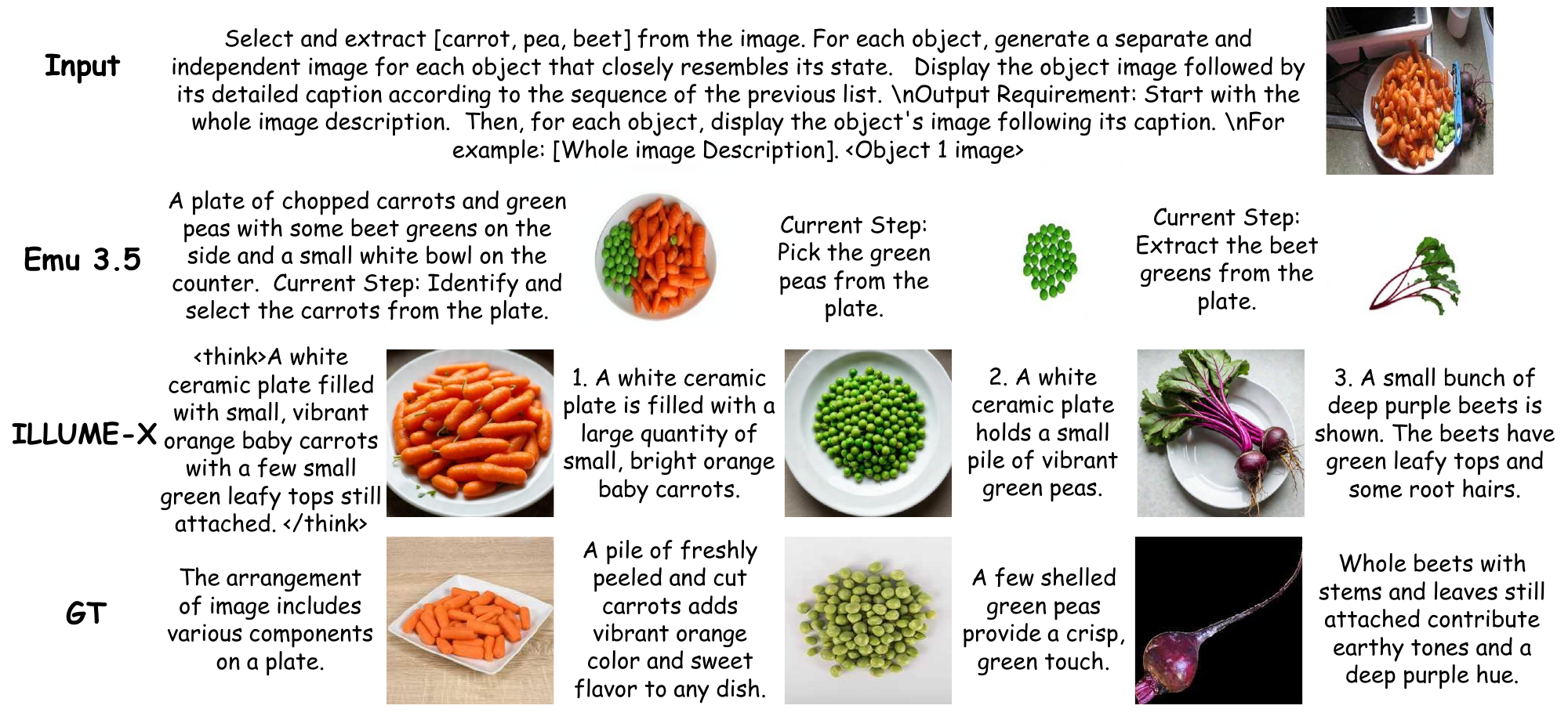}
    \caption{\textbf{Results of Image Decomposition}}
    \label{fig:Image_Decomposition}
\end{figure}

\begin{figure}
    \centering
    \includegraphics[width=\linewidth]{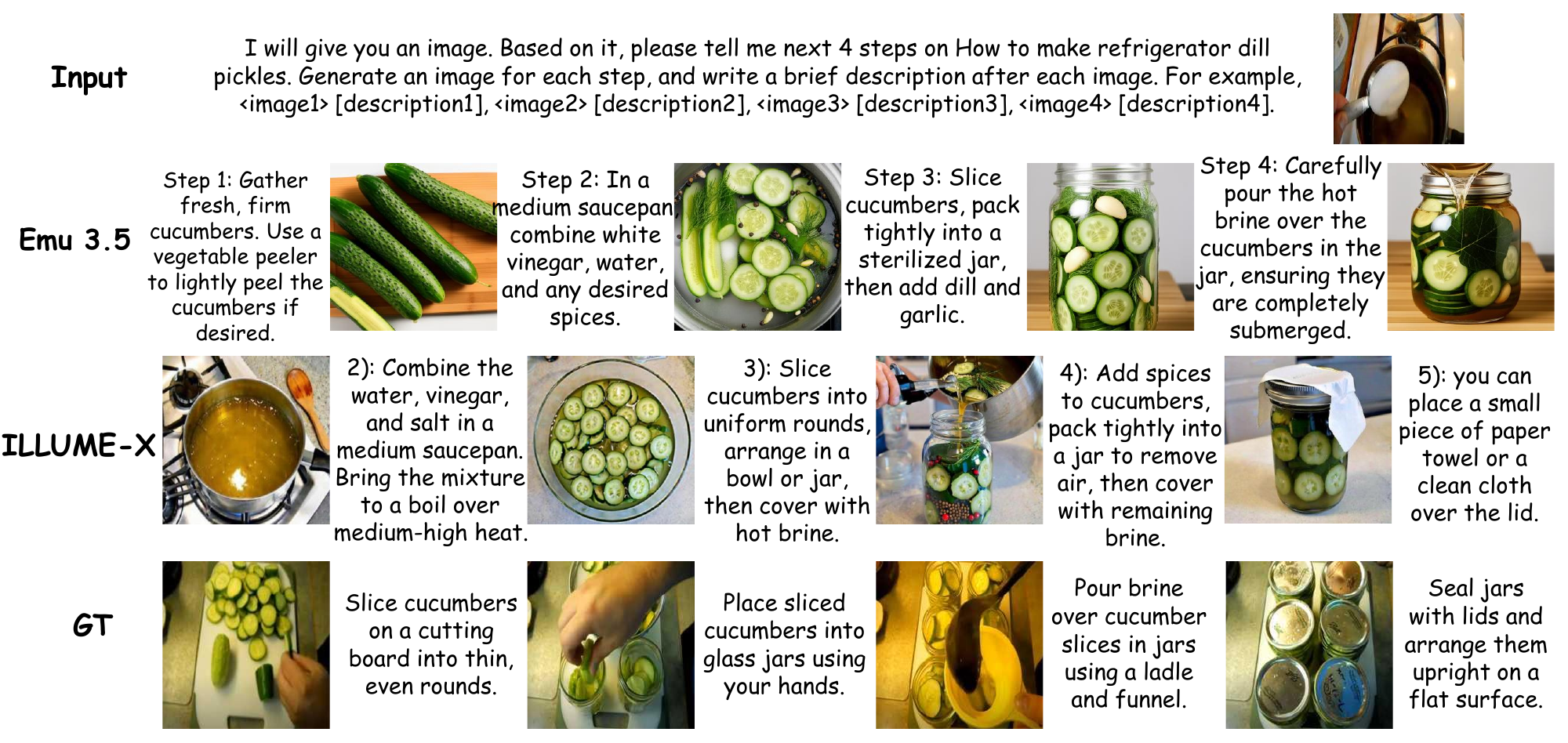}
    \caption{\textbf{Results of Image-Text Complementation}}
    \label{fig:Image-Text-Complementation}
\end{figure}

\begin{figure}
    \centering
    \includegraphics[width=\linewidth]{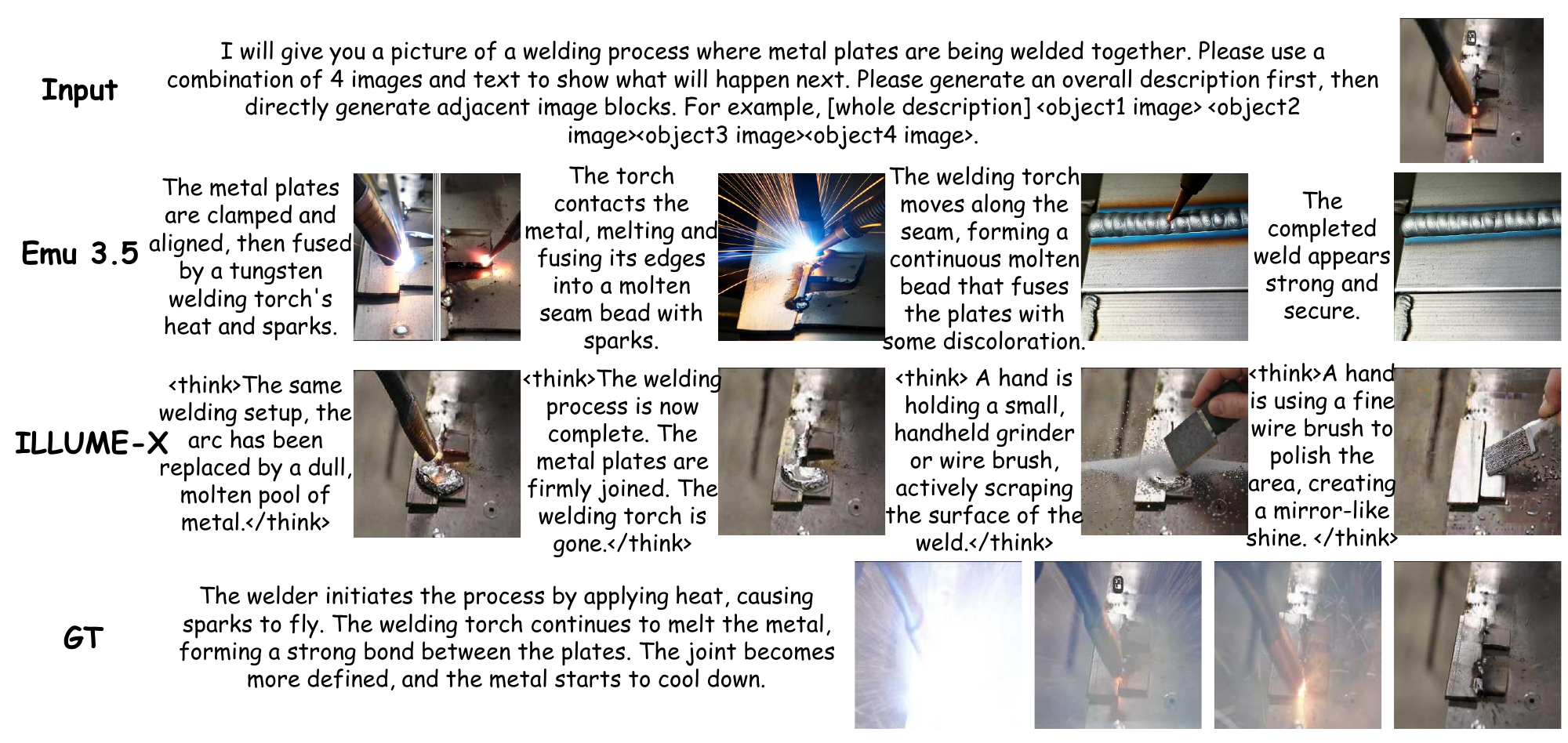}
    \caption{\textbf{Results of Temporal Prediction}}
    \label{fig:Temporal-Prediction}
\end{figure}

\begin{figure}
    \centering
    \includegraphics[width=\linewidth]{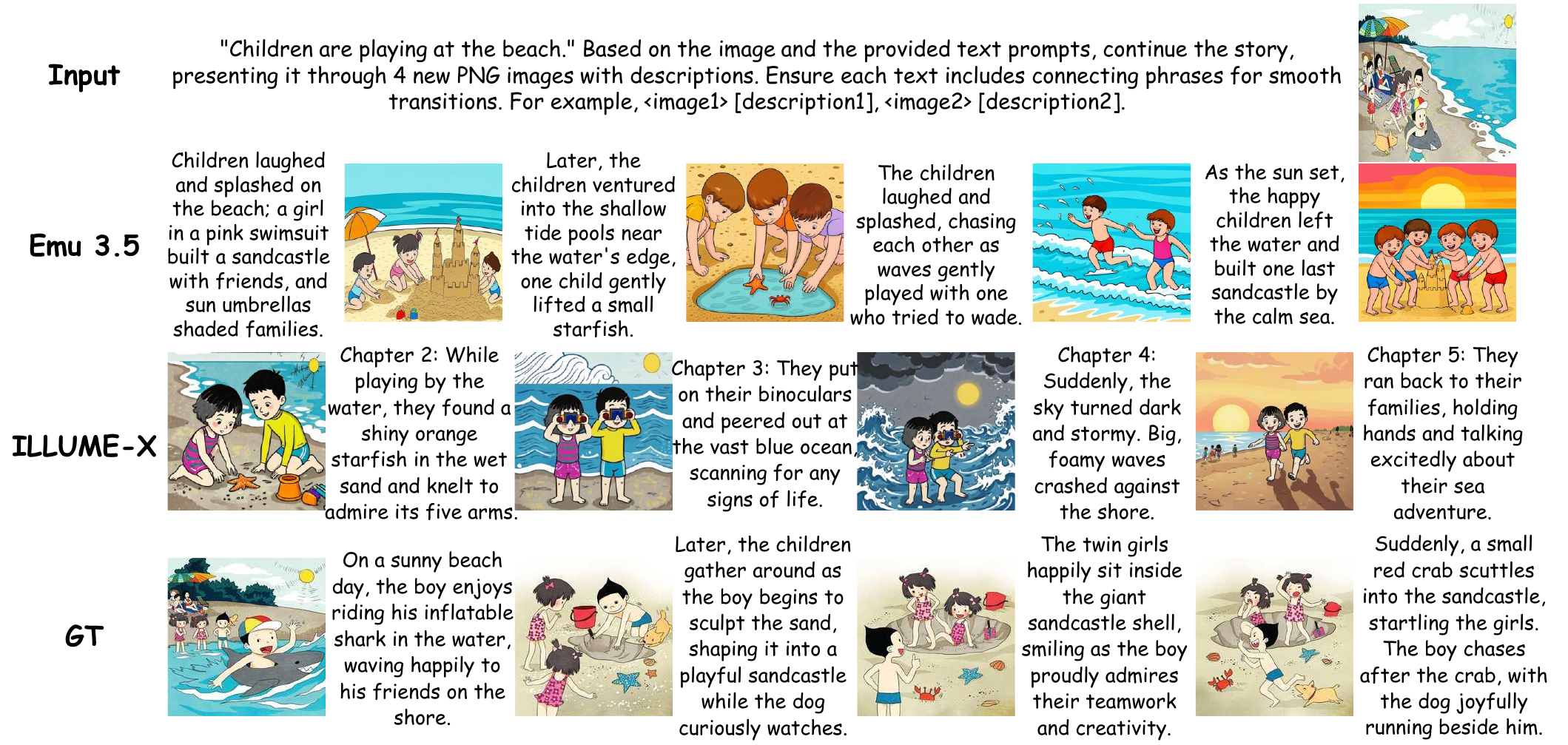}
    \caption{\textbf{Results of Visual Story Telling}}
    \label{fig:Visual-Story-Telling}
\end{figure}

\begin{figure}
    \centering
    \includegraphics[width=\linewidth]{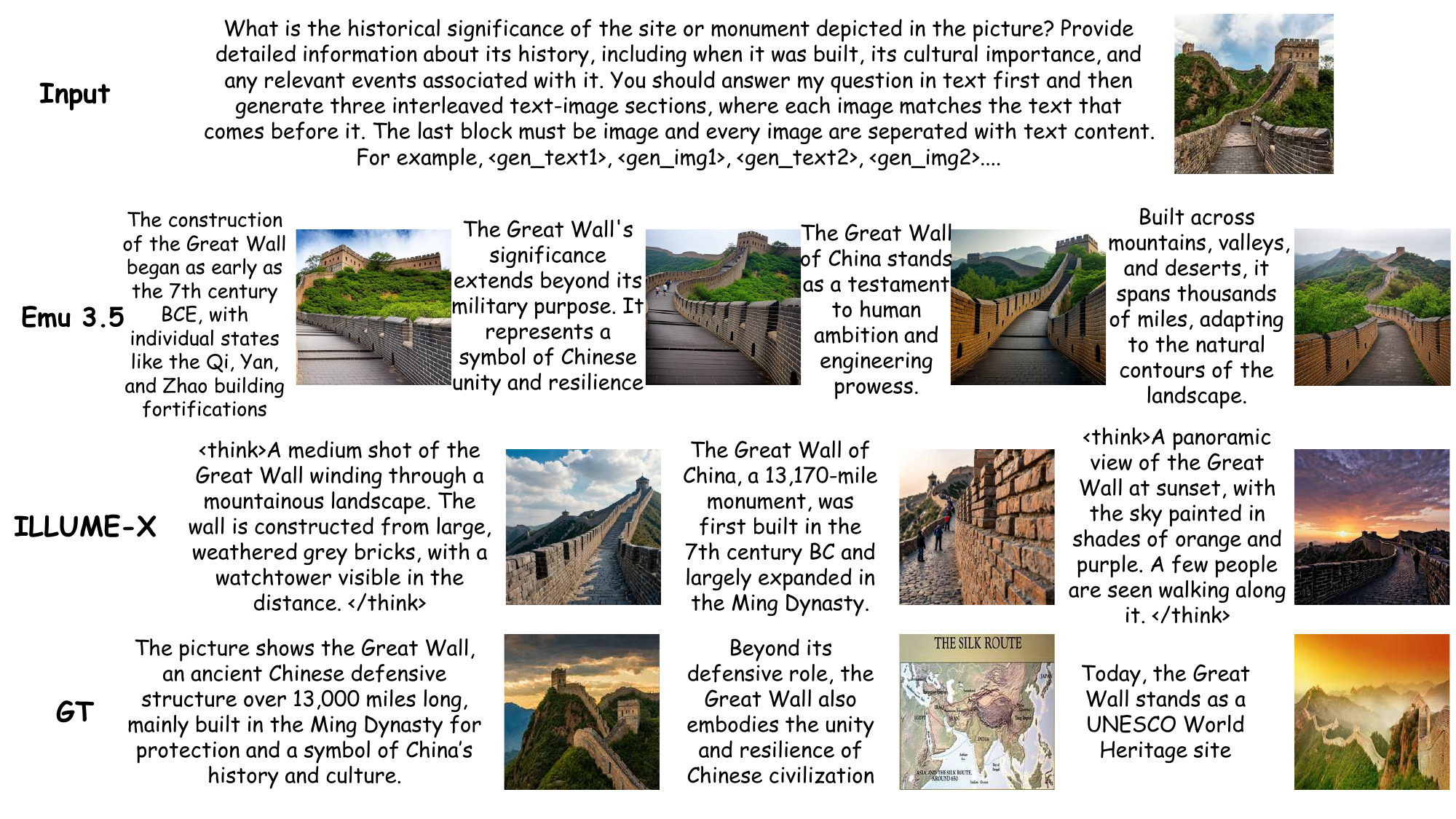}
    \caption{\textbf{Results of VQA}}
    \label{fig:vqa-result}
\end{figure}

We compared the experimental results more intuitively on the six major tasks of Style transfer, Progressive, 3D Scene, Image Decomposition, Image-Text Complementation, Temporal Prediction, Visual Story Telling and VQA, as shown in Fig.\ref{fig:style-transfer} to \ref{fig:vqa-result}.

\section{Custom Dataset for Interleaved}
\label{our-datasets}
\begin{figure}
    \centering
    \includegraphics[width=\linewidth]{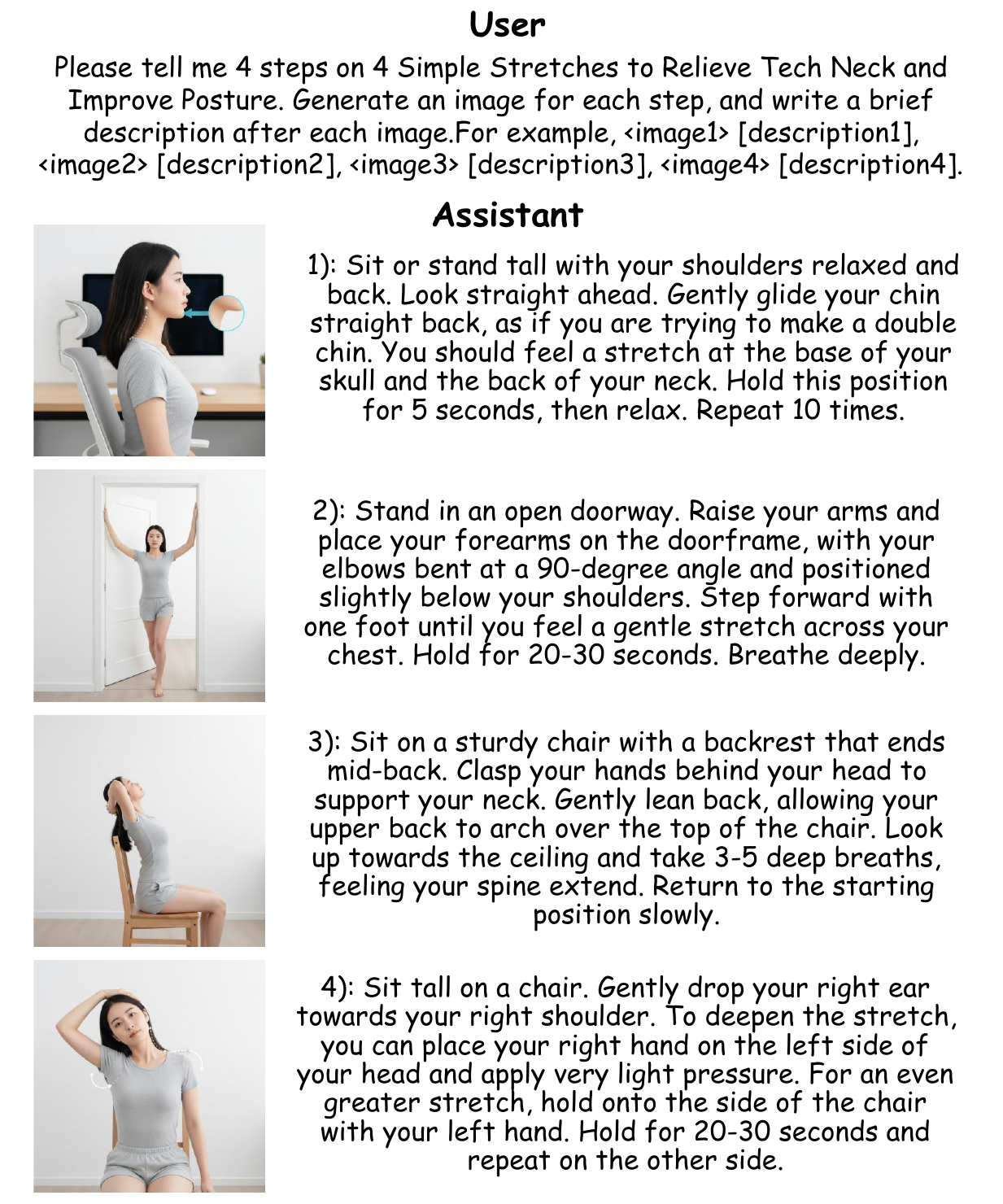}
    \caption{\textbf{Custom Dataset - Example 1}}
    \label{fig:our-howto}
\end{figure}

\begin{figure}
    \centering
    \includegraphics[width=\linewidth]{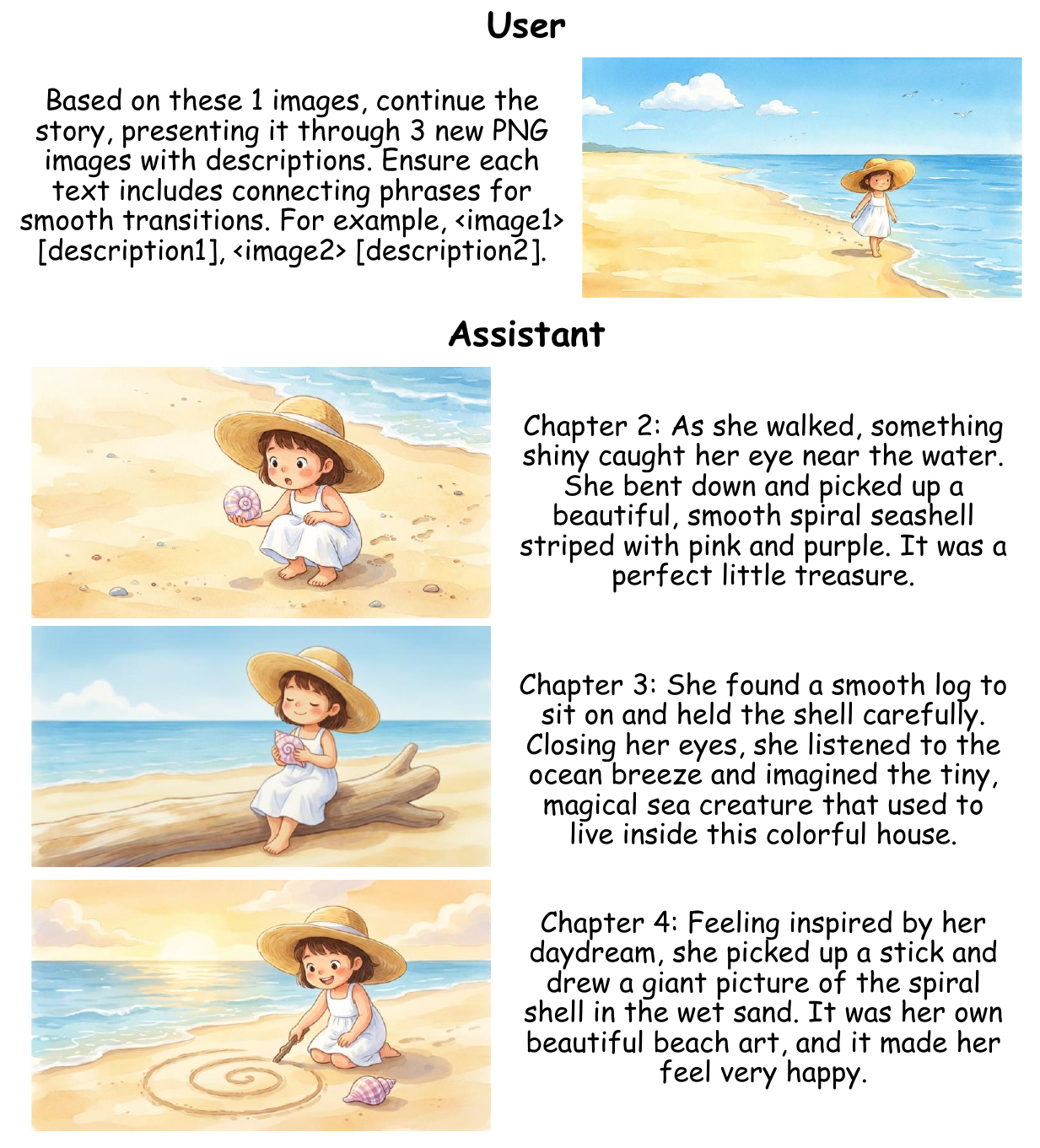}
    \caption{\textbf{Custom Dataset - Example 2}}
    \label{fig:our-story}
\end{figure}

\begin{figure}
    \centering
    \includegraphics[width=\linewidth]{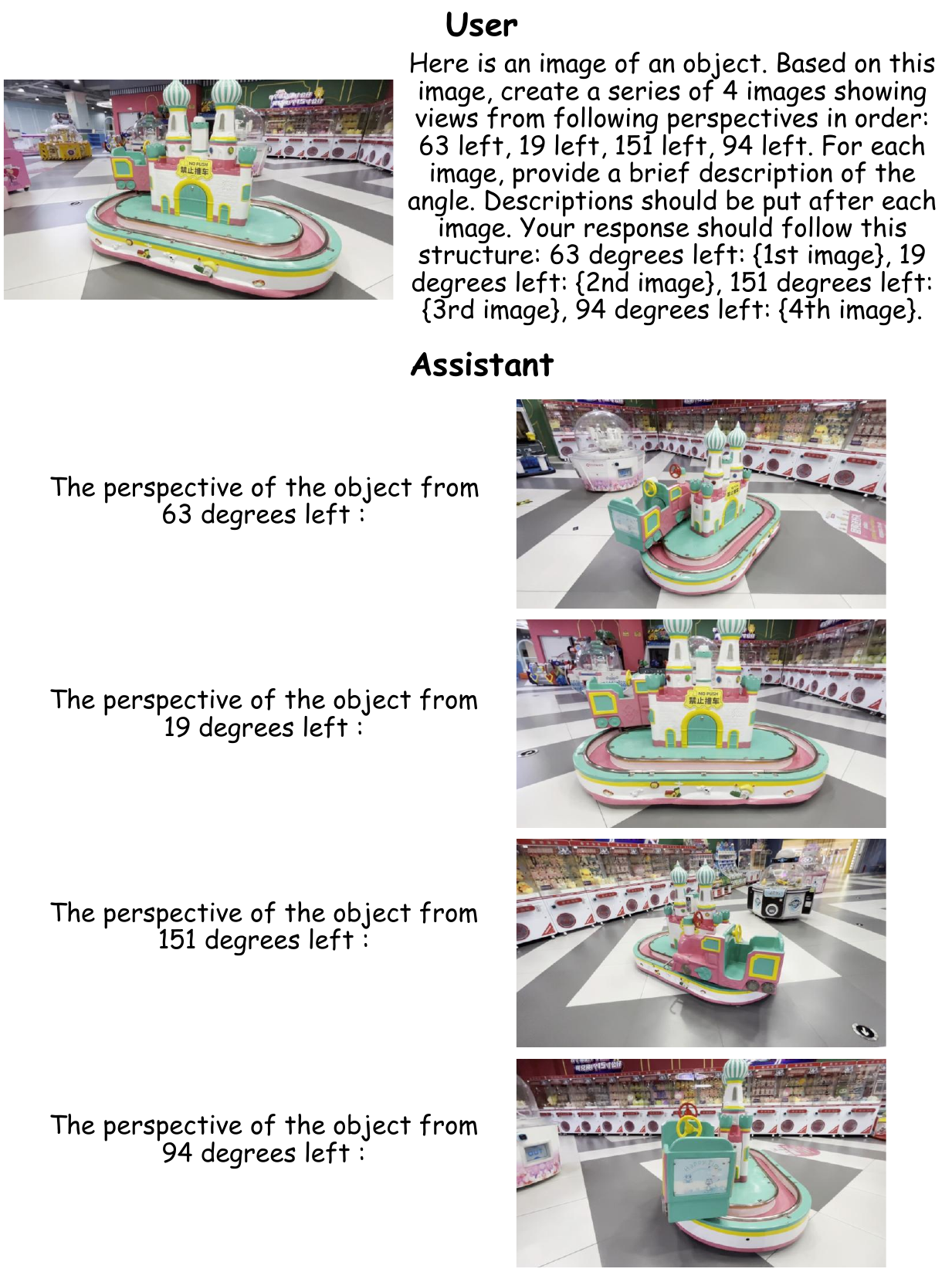}
    \caption{\textbf{Custom Dataset - Example 3}}
    \label{fig:our-3d}
\end{figure}

\begin{figure}
    \centering
    \includegraphics[width=\linewidth]{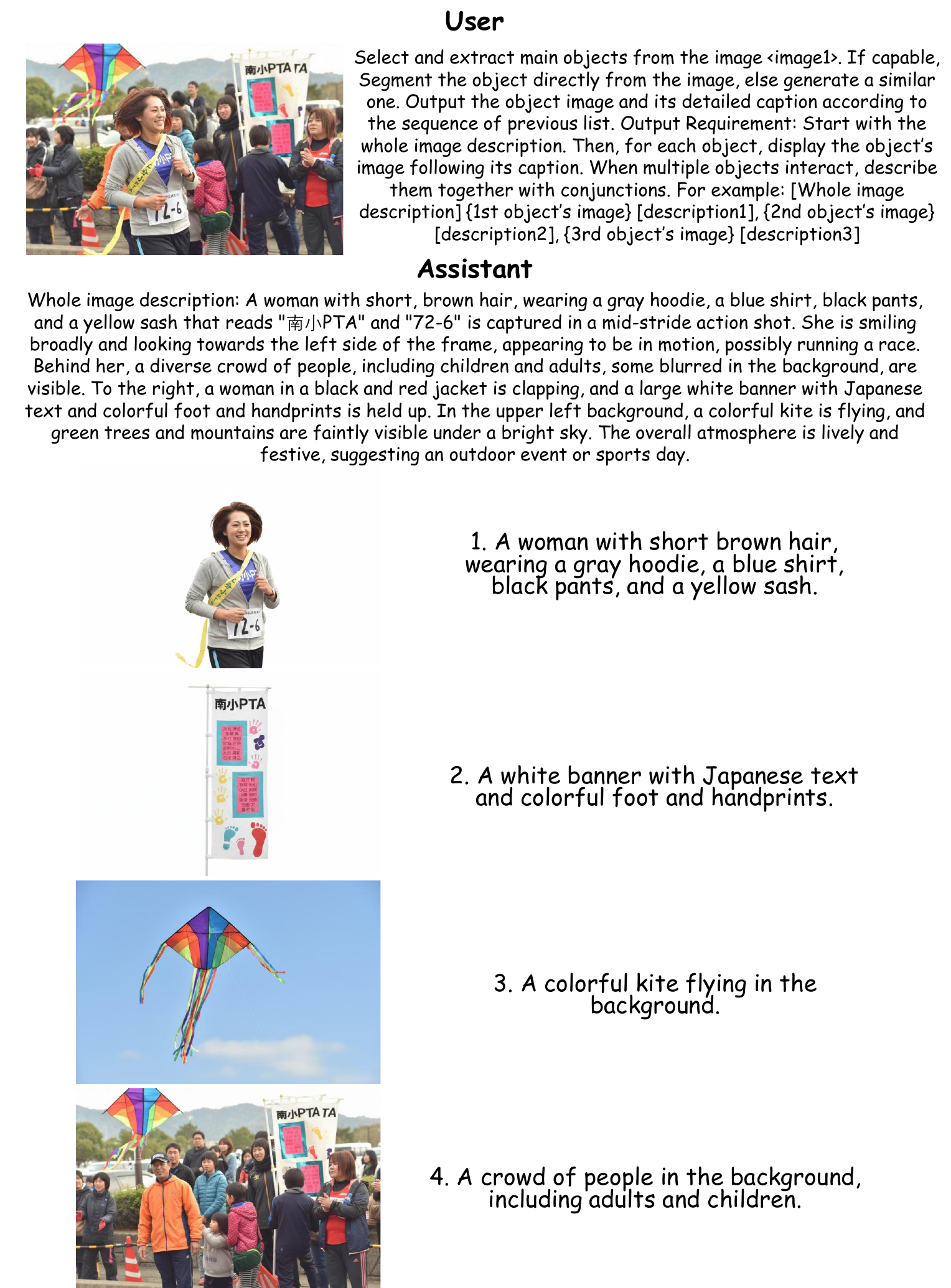}
    \caption{\textbf{Custom Dataset - Example 4}}
    \label{fig:our-decomposed}
\end{figure}

\begin{figure}
    \centering
    \includegraphics[width=\linewidth]{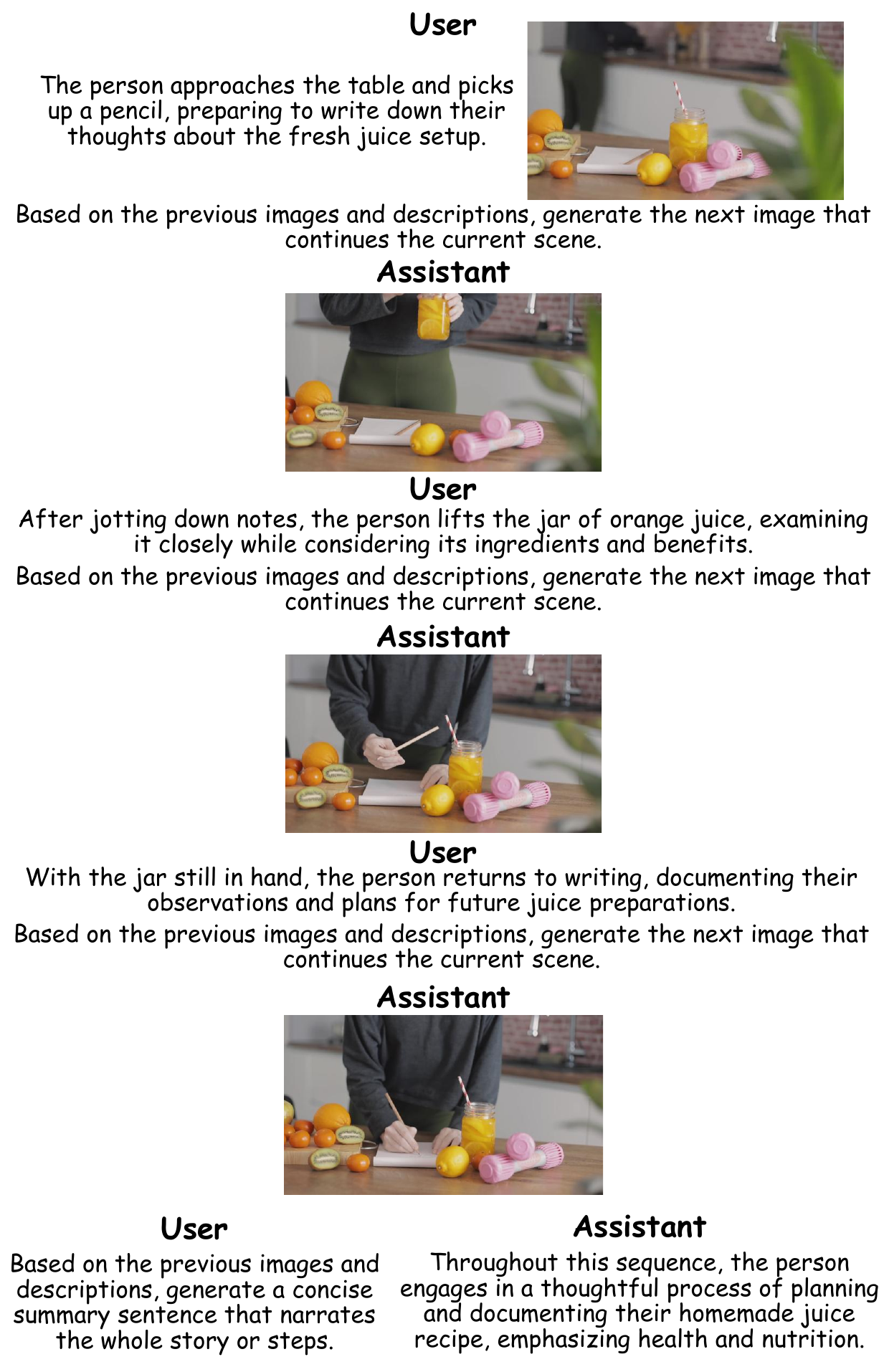}
    \caption{\textbf{Custom Dataset - Example 5}}
    \label{fig:our-vedio}
\end{figure}

We detail the pipeline for constructing our custom dataset in Sec.~\ref{Data_curation} of the main text. Here, we present several representative examples, as shown in Fig.~\ref{fig:our-howto} to~\ref{fig:our-vedio}.

\end{appendix}
\end{document}